\documentclass{article}

\usepackage[main,final]{neurips_2026}

\usepackage[utf8]{inputenc} 
\usepackage[T1]{fontenc}    
\usepackage{hyperref}       
\usepackage{url}            
\usepackage{booktabs}       
\usepackage{amsfonts}       
\usepackage{nicefrac}       
\usepackage{microtype}      
\usepackage{xcolor}         

\usepackage{microtype}
\usepackage{graphicx}
\usepackage{subcaption}
\usepackage{tabularx} 
\usepackage{array} 
\usepackage{multirow} 
\usepackage{siunitx} 
\usepackage{colortbl} 
\usepackage{amsmath}
\usepackage{amssymb}
\usepackage{mathtools}
\usepackage{amsthm}
\usepackage{enumitem}
\usepackage{xspace}
\usepackage{marvosym} 
\usepackage[capitalize,noabbrev]{cleveref}

\newcommand{\method}{\textsc{OrangeTree}\xspace}

\title{\method{}: A Linear and Tree-based Time Series Forecasting Model Supporting Multiple Input and Output Lengths}
\author{%
  {\normalfont\small\bfseries
  Yiqi Tang\textsuperscript{1,2},
  Zhichen Lai\textsuperscript{3},
  Yuxuan Yang\textsuperscript{1},
  Dalin Zhang\textsuperscript{4},
  Jinpeng Chen\textsuperscript{5},
  Gang Chen\textsuperscript{1},
  Huan Li\textsuperscript{1,2\Letter}} \\[0.5em]
  {\normalfont\small\textsuperscript{1}State Key Laboratory of Blockchain and Data Security, Zhejiang University} \\
  {\normalfont\small\textsuperscript{2}Hangzhou High-Tech Zone (Binjiang) Institute of Blockchain and Data Security} \\
  {\normalfont\small\textsuperscript{3}College of Computer and Data Science, Fuzhou University} \\
  {\normalfont\small\textsuperscript{4}Space Information Research Institute, Hangzhou Dianzi University} \\
  {\normalfont\small\textsuperscript{5}School of Computer Science, Beijing University of Posts and Telecommunications} \\[0.5em]
  {\normalfont\small\texttt{\{tang\_yiqi, yangyuxuan, cg, lihuan.cs\}@zju.edu.cn}} \\
  {\normalfont\small\texttt{lzc@fzu.edu.cn, zhangdalin@hdu.edu.cn, jpchen@bupt.edu.cn}}
}

\begin{document}

\maketitle

\begin{abstract}
Time series forecasting (TSF) models suffer from a critical rigidity: the inability to handle Multiple Input and Output Lengths (MIOL) within a single architecture. Current workarounds, such as padding or autoregression, often incur computational redundancy or suffer from error accumulation. We propose \method{}, a linear and tree-based architecture that decouples model parameters from sequence lengths. It employs a Segment Tree Encoder and an Inverse Tree Decoder to translate multi-length input history into multi-length output predictions within a unified framework. Crucially, a Range Weighter and a Feature Fuser bridge these components, dynamically selecting and fusing relevant ranges into optimal contexts for forecasting. Experiments on six benchmarks confirm \method{} achieves the SOTA performance. In MIOL settings, it matches the accuracy of length-fixed models without substantial accuracy degradation or a substantial increase in inference latency typical of heuristic methods. Code is available at \url{https://github.com/Tang-Yiqi/OrangeTree}.
\end{abstract} 

\section{Introduction}
\label{sec:intro}
Time series forecasting (TSF) is a cornerstone of modern decision systems, underpinning critical domains like energy, transportation, and finance~\cite{benidis2022deep}. 
While the transition from statistical baselines to deep architectures~\cite{jiang2025multi}, ranging from Transformers-based~\cite{zhou2021informer,liu2022pyraformer} to Linear-based models~\cite{zeng2023transformers,das2023long}, has yielded substantial accuracy gains, the field remains entrenched in a paradigm of \emph{structural rigidity}.

\begin{figure}[htbp]
\centering
\includegraphics[width=0.9\textwidth]{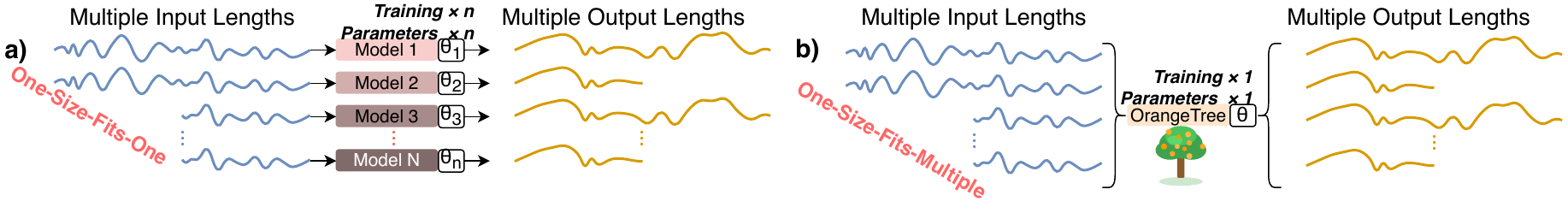}
\caption{Paradigm Shift in TSF: (a) The traditional rigid approach requires training distinct models for each input-output configuration. (b) The proposed \method{} paradigm enables a single model to natively support MIOL during inference.}
\label{fig:multi}
\end{figure}

The defining limitation of SOTA architectures is their inability to handle \textbf{Multiple Input and Output Lengths (MIOL)} natively. 
As shown in Figure~\ref{fig:multi}(a), the mainstream paradigm enforces a ``One-Size-Fits-One'' constraint: input and output lengths must be permanently fixed at initialization. This rigidity creates a severe friction between model design and real-world deployment. In dynamic environments (Figure~\ref{fig:multi}(b)), systems often need short contexts for fast inference, yet long contexts for accurate long-term planning. To bridge this gap, practitioners resort to inefficient ensembles of length-specific models, wasting storage and management overhead, or crude heuristics like padding and trimming~\cite{bier2022variable}, leading to computational redundancy and extraneous noise.

The root of this rigidity lies in the entanglement of model parameters with sequence lengths. Recurrent architectures, while theoretically flexible, suffer from error accumulation over long horizons~\cite{song2024deep}. Conversely, modern Linear and Transformer models use fixed-size embeddings and static prediction heads, tightly coupling parameters to sequence length.
To adapt these rigid structures to MIOL, external workarounds (such as padding and trimming~\cite{bier2022variable}) are required, but these lead to wasted computational resources or information loss (see our discussions in Section~\ref{ssec:related}).

In this work, we challenge this status quo by posing a pivotal question: \textit{Is it possible to design a TSF architecture that, through a single training process, supports multiple input and output lengths during inference?}

We answer affirmatively and propose \method{}, a linear and tree-based model that delivers native flexibility for TSF. Its core intuition is to view a time series as a set of ranges rather than a fixed sequence. \method{} derives its name from its ability to assess the importance of all ranges ($\text{All Range} \Rightarrow \text{Orange}$) and to perform representation extraction and prediction using a tree structure.

\method{} introduces a structural paradigm shift through four key mechanisms. First, we replace fixed linear projections with a \textbf{Segment Tree Encoder} grounded in \textit{binary lifting}. This allows the model to sparsely encode variable-length history into multi-scale representations by sharing aggregation weights, decoupling parameters from input length. Second, inspired by Pascal's Triangle, we introduce a \textbf{Range Weighter} that scores input range importance layer-by-layer. Then, the \textbf{Feature Fuser} orchestrates the Encoder by range importance for the extraction and fusion of the top-$K$ range representations. Finally, an \textbf{Inverse Tree Decoder} synthesizes the forecast by dynamically generating leaf nodes to match any desired output length.
Our extensive evaluation shows that \method{} does not merely solve the MIOL problem --- it does so without substantial performance degradation. To draw an analogy: in a running competition, athletes train for specific distances, just as a short‑distance runner usually cannot win a long‑distance race. In machine learning, a single parameter set usually struggles to adapt to diverse scenarios (MIOL). In this work, we train an "all-around athlete" (\method{} trained on MIOL) that not only rivals the "single-event champion" (\method{} trained on fixed lengths) trained for each length, but also outperforms existing length-specific SOTAs.

Our contributions are summarized as follows:
\begin{itemize}[noitemsep, topsep=0pt, leftmargin=*]
\item We propose \method{}, a novel tree-based forecasting framework that natively supports multiple input and output lengths (MIOL) without padding or re-training.

\item We design a Segment Tree Encoder and an Inverse Tree Decoder grounded in \textit{binary lifting}. Paired with a hierarchical Range Weighter and a top-$K$ Feature Fuser, these structures dynamically generate multi-granularity representations to construct optimal contexts for specified output lengths.

\item Extensive experiments confirm that \method{} establishes a new SOTA on standard benchmarks while natively resolves the MIOL challenge, without parameter inflation, computational redundancy, or severe performance degradation.
\end{itemize}

\section{Preliminaries}
\subsection{Related Work}
\label{ssec:related}
Time series forecasting has evolved from statistical methods to deep learning paradigms~\cite{jiang2025multi}. Recent advances primarily focus on forecasting accuracy and efficiency, but the ability to natively support MIOL remains rarely addressed. This section reviews representative methods and analyzes their structural limitations in multi-length modeling.

\textbf{Structural Rigidity in TSF Architectures.} With the proposal of the attention mechanism~\cite{vaswani2017attention}, a surge of Transformer-based TSF models have emerged. Early Transformer variants, such as Informer~\cite{zhou2021informer} and Pyraformer~\cite{liu2022pyraformer}, focused on reducing complexity in long-term dependency modeling through ProbSparse self-attention and hierarchical graph structures, respectively. Recent methods, like PatchTST~\cite{nie2023time} and iTransformer~\cite{liu2024itransformer}, shift the paradigm toward patch-based representation and channel-level attention to improve accuracy. During this period, the emergence of DLinear~\cite{zeng2023transformers}, an extremely simple linear model, completely broke the years-long ``monopoly'' of Transformer structures in TSF, proving that Linear models are also capable of TSF tasks. Subsequently, numerous Linear TSF models have emerged, such as TiDE~\cite{das2023long}, TSMixer~\cite{ekambaram2023tsmixer}, SOFTS~\cite{han2024softs}, etc., improving interactions between time steps and channels. Therefore, structurally, modern TSF models can be divided into two main branches: Transformer models and Linear models.

Although Transformer and Linear models in modern TSF differ sharply in structure, they share a trait: modeling length is highly coupled with model parameters. Existing Linear models essentially rely on linear transformations like \( \mathbf{Y} = \mathbf{W}\mathbf{X} + \mathbf{B} \), where the dimensions of the weight matrix \( \mathbf{W} \) and bias \( \mathbf{B} \) are hard-coded to specific output lengths. Similarly, Transformer models typically depend on fixed-length positional encodings and static embedding layers on the input side, while employing single or multi-layer linear forecasting heads on the output side. Furthermore, some niche architectures, such as CNN-based (e.g., TCN~\cite{bai2018empirical}) and State-Space-based (e.g., S-Mamba~\cite{wang2025mamba}), also lose the flexibility to vary the modeling length due to the necessary linear transformations on their input and output sides. Consequently, most modern TSF models cannot natively achieve MIOL.

\textbf{Multiple-Length Modeling in Neural Networks.} Although variable-length modeling capabilities are well-established in the NLP and CV, transferring these paradigms to the TSF domain remains challenging. In NLP, variable-length modeling is achieved via autoregressive Transformer structures~\cite{vaswani2017attention}. However, for continuous time series, this method leads to significant cumulative errors and delays compared to non-autoregressive approaches. Similarly, in CV, dynamic resolution image modeling is often facilitated by AdaPool~\cite{stergiou2022adapool}. However, this approach may disrupt the complex structures of time series and dependencies among channels.

Thus, TSF lacks a dedicated solution for MIOL. Early methods like RNN~\cite{elman1990finding} and LSTM~\cite{hochreiter1997long} theoretically support MIOL through recurrent connections. However, inherent issues such as difficulty in modeling long-term dependencies and error accumulation~\cite{song2024deep} cause these models to lag far behind modern TSF models in forecasting accuracy. To circumvent structural rigidity, existing methods employ heuristic adapters like Padding and Autoregression~\cite{bier2022variable}, but these all result in varying degrees of resource waste or performance loss. Recent methods, such as HDMixer~\cite{huang2024hdmixer}, use linear interpolation before input modeling, theoretically supporting multiple input lengths, but still employ fixed forecasting heads on the output side. Additionally, some Foundation Model-based methods theoretically support MIOL, like LLMTIME~\cite{gruver2023large}, LLM4DyG~\cite{zhang2024llm4dyg}, and LSTPrompt~\cite{liu2024lstprompt}. However, these methods incur high computational cost and latency.

Unlike these compromise methods, \method{} introduces a linear and tree-based architecture that natively decouples parameters from sequence length, providing the first end-to-end and unified MIOL solution for the TSF field.

\subsection{Problem Statement}
\label{ssec:problem}

Formally, let $\mathbf{X} = \{\mathbf{x}_1, \mathbf{x}_2, \dots, \mathbf{x}_N\}$ denote a multivariate time series, where $\mathbf{x}_t \in \mathbb{R}^{C}$ represents the observations of $C$ variables (channels) at time step $t$.
Given a historical look-back window of length $L$, denoted as $\mathbf{X}_{in} = [\mathbf{x}_{t-L+1}, \dots, \mathbf{x}_t] \in \mathbb{R}^{L \times C}$, the canonical TSF task aims to predict the subsequent forecast horizon of length $T$, denoted as $\mathbf{X}_{out} = [\mathbf{x}_{t+1}, \dots, \mathbf{x}_{t+T}] \in \mathbb{R}^{T \times C}$.

\textbf{The MIOL Objective.}
Unlike standard TSF formulations where $L$ and $T$ are fixed hyperparameters binding the model architecture, the MIOL task requires learning a unified mapping $f_{\theta}: \mathbb{R}^{L \times C} \to \mathbb{R}^{T \times C}$. Crucially, the parameter set $\theta$ must be invariant to the lengths, i.e., $\dim(\theta) \perp \{L, T\}$.

In this work, we define the MIOL objective over a discrete set of valid configurations $\mathcal{S} = \{(L_n, T_n)\}_{n=1}^N$. The model is trained to minimize the cumulative error across all pairs in $\mathcal{S}$ and then supports inference for any $(L, T) \in \mathcal{S}$ without retraining. We adopt this discrete strategy for two key reasons: 1) in real-world operational environments (e.g., cloud-based forecasting services), varying demands typically cluster around specific, standardized horizons (e.g., short-term vs. long-term), thus discrete configurations cover the vast majority of real-world needs; 2) learning dependencies over a continuous length spectrum introduces excessive variance, making convergence difficult. Restricting the domain to a discrete set $\mathcal{S}$ ensures peak performance on every supported configuration.

\section{The \method{} Design}
\label{sec:method}
\begin{figure*}[ht]
\centering
\includegraphics[width=1\textwidth]{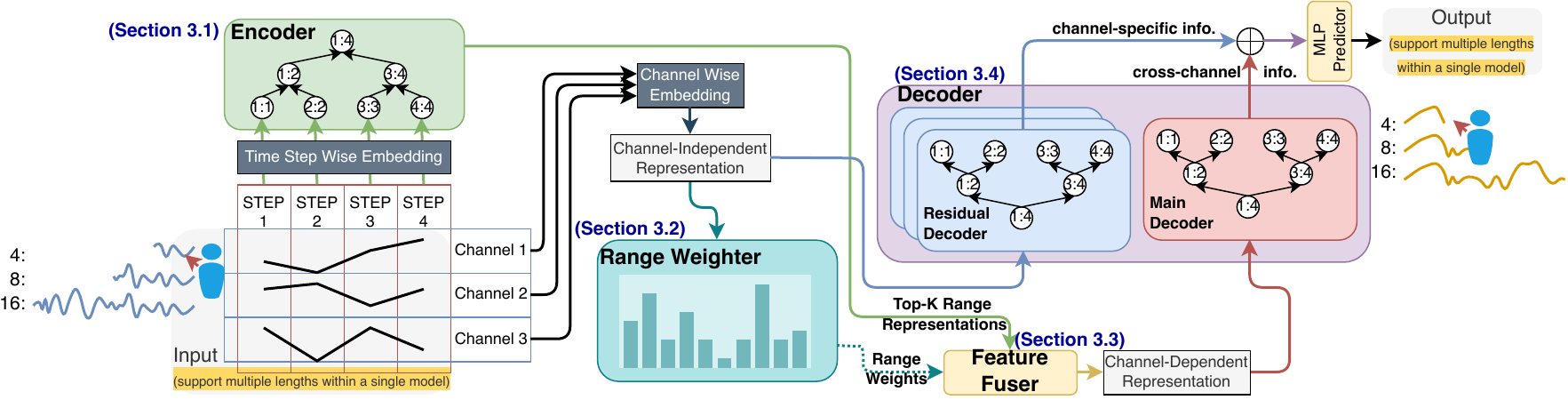}
\caption{\method{} architecture with a toy example ($C=3, L=4$ for an input): Encoder aggregates time steps into range representations (e.g., 1-2, 2-4, and 1-4) using a linear layer; Range Weighter and Feature Fuser compute importance across ranges and fuse the most informative features; Decoder produces variable-length forecasts by expanding parallel Main and Residual inverse trees.}
\label{fig:simplified}
\end{figure*}

Figure~\ref{fig:simplified} illustrates the \method{} architecture, designed to decouple model parameters from sequence lengths. The framework processes multivariate time series through four linear modules:
\begin{enumerate}[noitemsep, topsep=0pt, leftmargin=*]
\item \textbf{\emph{Segment Tree Encoder}} (Section~\ref{ssec:segment_tree_encoder}) for representation: A hierarchical encoder that projects the input series into a multi-scale tree structure. By employing shared linear layers across nodes, it extracts channel-dependent embeddings for \textit{each} possible sub-range $[i:j]$ of the historical window, independent of the input length $L$.

\item \textbf{\emph{Range Weighter}} (Section~\ref{ssec:range_weighter}) for importance:
A global context evaluator inspired by Pascal’s Triangle. It operates on a channel-independent view to score the importance  across time series ranges, acting as a linear attention mechanism to identify the most predictive historical ranges.

\item \textbf{\emph{Feature Fuser}} (Section~\ref{ssec:feature_fuser}) for selection and fusion:
An aggregation module bridging the Segment Tree Encoder and Range Weighter. It uses the importance scores to select the most relevant ranges from the Encoder's tree and fuses them into a compact, channel-dependent hidden state.

\item \textbf{\emph{Inverse Tree Decoder}} (Section~\ref{ssec:inverse_tree_decoder}) for prediction:
A dual-branch decoding architecture. It combines the channel-dependent (Main Decoder) and channel-independent (Residual Decoder) representations via an inverted tree, dynamically expanding depth to match the output length $T$.
\end{enumerate}

The key innovation enabling MIOL is the \textit{binary lifting} nature of both the Segment and Inverse Trees. Since the linear parameters ($\theta$) are defined at the node level rather than the sequence level, \method{} can adapt its depth and breadth for any given $(L, T)$ pairs during inference.

\subsection{Segment Tree Encoder}
\label{ssec:segment_tree_encoder}

To address the rigidity of fixed-size embedding layers (where $\mathbf{W} \in \mathbb{R}^{D \times L}$), we propose a hierarchical architecture inspired by the classic segment tree data structure. Unlike standard linear models that treat the input history as a flat vector, our encoder projects the time series into a recursive tree structure. This design yields two critical properties: (1) \emph{Length Invariance}, as parameters are shared across all tree nodes independent of the total sequence length $L$; and (2) \emph{Multi-Scale Representation}, enabling the efficient extraction of representations for any range $[i:j]$ in $\mathcal{O}(\log L)$ time.

\begin{figure*}[ht]
\centering
\includegraphics[width=1\textwidth]{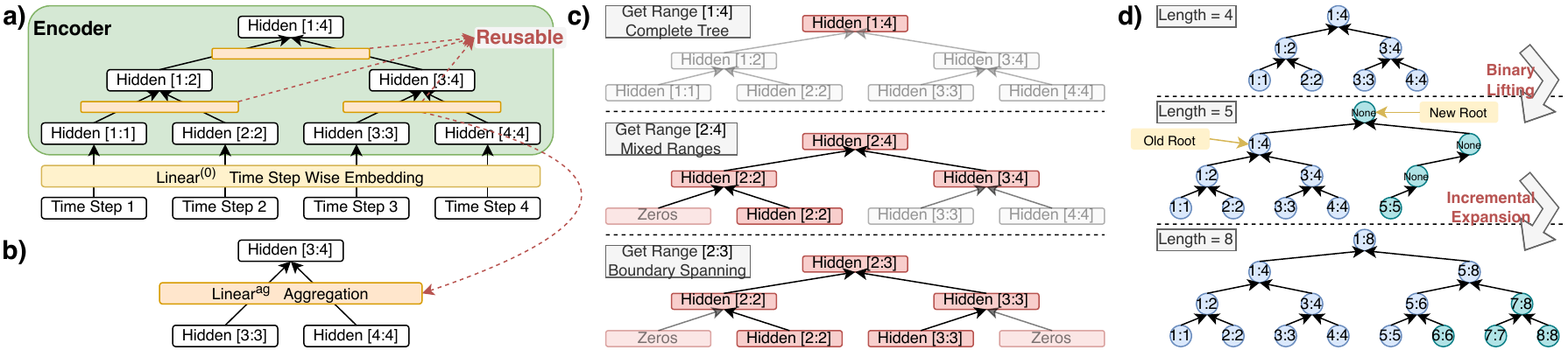}
\caption{Mechanics of the Segment Tree Encoder. a \& b) The encoder recursively aggregates time steps using a shared linear layer, ensuring length-independent parameters; c) Arbitrary ranges' representations like $[2:4]$ are efficiently extracted by stitching together the minimal set of pre-computed nodes; d) The tree dynamically expands via \textit{binary lifting} to fit variable input lengths.}
\label{fig:encoder}
\end{figure*}

\textbf{Recursive Representation Extraction.}
Given a multivariate input $\mathbf{X} = \{x_1, \dots, x_L\}$ with $x_t \in \mathbb{R}^C$, the encoder builds the tree bottom-up. First, each time step is linearly projected into leaf nodes:
\begin{equation}
{\mathbf{h}}_{t} = \operatorname{Linear}^{(0)}\left( {\mathbf{x}}_{t}\right)  \in  {\mathbb{R}}^{D}, t = 1 \sim  L.
\end{equation}
Higher-level nodes are then recursively aggregated from their children. For a parent node covering the range $[i:j]$, its representation $\mathbf{h}_{[i:j]}$ is derived from its left child $\mathbf{h}_{\text{left}}$ and right child $\mathbf{h}_{\text{right}}$ via a shared aggregation layer:
\begin{equation}
\label{eq:agg}
{\mathbf{h}}_{\text{parent}} = \operatorname{Linear}^{ag}\left( \left\lbrack  {{\mathbf{h}}_{\text{left}};{\mathbf{h}}_{\text{right}}}\right\rbrack  \right)  \in  {\mathbb{R}}^{D},
\end{equation}
where $[\cdot ; \cdot]$ denotes concatenation. Crucially, because $\mathbf{h}_{\text{parent}}$ maintains the same dimension $D$ as its children, the $\operatorname{Linear}^{ag}$ in Eq.~\ref{eq:agg} is reused across all layers of the tree. This parameter sharing is the mechanism that decouples the model size from the input length $L$ (see Figure~\ref{fig:encoder}(b)). 

To represent an arbitrary range $[u:v]$, \method{} identifies its \emph{canonical covering set}, the minimal collection of disjoint, pre-computed nodes spanning the range. Figure~\ref{fig:encoder}(c) illustrates three retrieval patterns: direct access for complete subtrees (e.g., $[1:4]$ via the root), hierarchical composition for mixed ranges (e.g., $[2:4]$ via leaf $[2:2]$ and node $[3:4]$), and boundary-spanning aggregation (e.g., $[2:3]$ via leaves $[2:2]$ and $[3:3]$). Details in Appendix~\ref{sssec:RepExt}.

\textbf{Dynamic Expansion via \textit{Binary Lifting}.}
A core challenge in MIOL is handling $L$ without retraining. We solve this via a \emph{Binary Lifting} strategy. The tree is not initialized with a fixed depth; instead, it dynamically expands to the smallest depth $l$ such that $2^l \ge L$. As shown in Figure~\ref{fig:encoder}(d), when the input length increases (e.g., from $L=4$ to $L=5$), the model simply adds new nodes and potentially a new root, reusing the existing $\operatorname{Linear}^{ag}$ for the new connections. This ensures the model capacity scales naturally with input length while maintaining $\mathcal{O}(L)$ space complexity.

\textbf{Generalization to Multi-Branch Trees.}
While binary trees are efficient, deep trees can suffer from optimization difficulties (e.g., vanishing gradients). To mitigate this, we generalize the encoder to a $B$-branch segment tree, where each parent aggregates $B$ children. This introduces a trade-off: larger $B$ reduces tree depth ($\log_B L$) and facilitates gradient flow, but increases the parameter count of the aggregation layer ($\mathbb{R}^{B \cdot D} \to \mathbb{R}^D$). We treat $B$ as a structural hyperparameter to balance expressivity and efficiency. See the hyperparameter settings for $B$ in Appendix~\ref{ssec:hyper}.

\subsection{Range Weighter}
\label{ssec:range_weighter}

\begin{figure}[ht]
\centering
\includegraphics[width=0.9\textwidth]{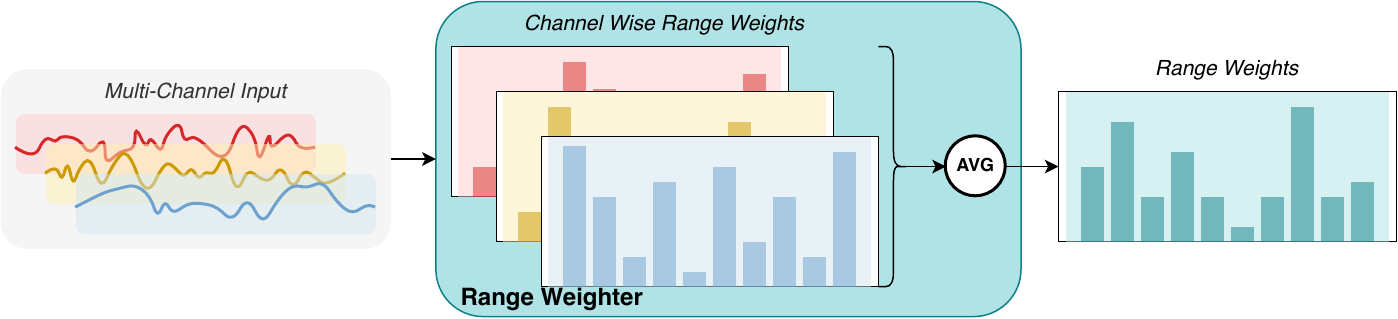}
\caption{Overview of the Range Weighter.}
\label{fig:weighter}
\end{figure}

To distinguish informative historical patterns from noise, we introduce a global attention mechanism that evaluates the predictive importance across ranges $[i:j]$ within the input window. As shown in Figure~\ref{fig:weighter}, this weighter employs a resolution normalization strategy to decouple the scoring parameters from the input length $L$, and a hierarchical structure inspired by Pascal's Triangle to efficiently propagate global context to local ranges.

\textbf{Resolution Normalization via Interpolation.}
To ensure the weighting is invariant to $L$, we first project the input $\mathbf{X} \in \mathbb{R}^{L \times C}$ into a fixed-resolution hidden space. We sample $M$ fixed points $\{\tilde{\mathbf{x}}_1, \ldots, \tilde{\mathbf{x}}_M\}$ via linear interpolation, where $M$ is a structural hyperparameter. Each channel is then converted to a channel-independent embedding:
\begin{equation}
\label{equ:e_c}
\mathbf{e}_c = \operatorname{Linear}^{\text{RM}}(\tilde{\mathbf{x}}_{:,c}) \in \mathbb{R}^D, \quad c = 1 \sim C,
\end{equation}
where $\tilde{\mathbf{x}}_{:,c} \in \mathbb{R}^M$ denotes the interpolated series of channel $c$. The vector $\mathbf{e}_c$ encapsulates the global representation of the channel and serves as the channel-independent representation in Decoder (see Residual Decoder in Section~\ref{ssec:ResidualDecoder}).

\textbf{Root Node Initialization.}
Since it only scores importance, we project the high-dimensional channel embedding $\mathbf{e}_c$ into a lightweight hidden space $D' \ll D$:
\begin{equation}
\mathbf{g}_{[1:L]}^{(c)} = \operatorname{Linear}^{\text{root}}(\mathbf{e}_c) \in \mathbb{R}^{D'},
\end{equation}
which then acts as the root node of the Pascal structure, representing the hidden state of the full input range $[1:L]$.

\textbf{Pascal-Style Hierarchical Scoring.}
A top-down generation process inspired by Pascal's Triangle derives weights for all ranges. From $[1:L]$, we recursively generate hidden states for smaller ranges by ``shaving off'' time steps from the boundaries. Specifically, the representation for $[i:j]$ is aggregated from its two ``parent'' ranges, $[i-1:j]$ (extended left) and $[i:j+1]$ (extended right):
\begin{equation}
\mathbf{g}_{[i:j]}^{(c)} = \operatorname{Linear}^{\text{pascal}}([\mathbf{g}_{[i-1:j]}^{(c)}; \mathbf{g}_{[i:j+1]}^{(c)}]) \in \mathbb{R}^{D'}.
\end{equation}
This recursive formula allows for assessing a local range's importance relative to its broader context. 
Finally, range scalar scores are averaged across channels to obtain a unified global weight:
\begin{equation}
w_{[i:j]} = \frac{1}{C}\sum_{c=1}^{C} w_{[i:j]}^{(c)}, w_{[i:j]}^{(c)} = \operatorname{Linear}^{\text{vote}}(\mathbf{g}_{[i:j]}^{(c)}).
\end{equation}
An example is provided in Appendix~\ref{sssec:WeightCalc}. These weights are subsequently used by the {Feature Fuser} to select the most predictive ranges from the Encoder.

\subsection{Feature Fuser}
\label{ssec:feature_fuser}
\begin{figure}[ht]
\centering
\includegraphics[width=0.9\textwidth]{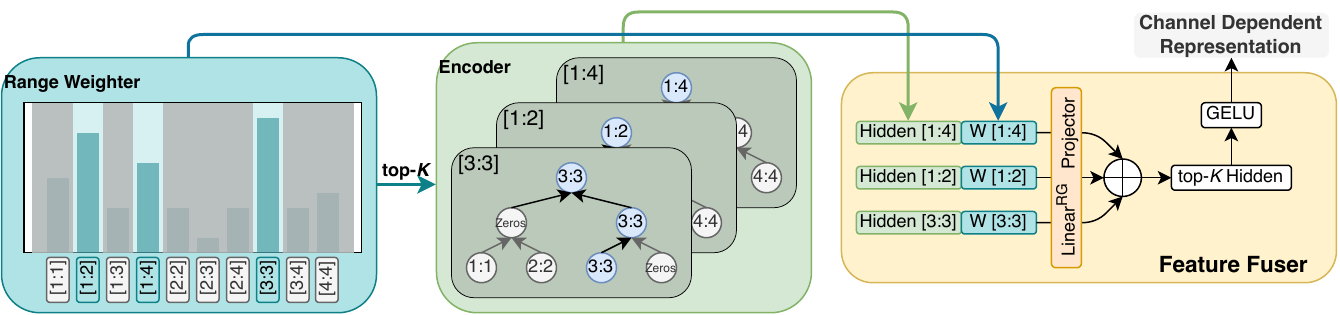}
\caption{Architecture of the Feature Fuser, combining the Range Weighter and Encoder to produce a channel‑dependent representation.}
\label{fig:fuser}
\end{figure}

The fuser serves as a sparse filter, aggregating the most predictive historical ranges. As illustrated in Figure~\ref{fig:fuser}, guided by the importance weights $\{w_{[i:j]}\}$ from the Range Weighter, it selects the top-$K$ (see the hyperparameter settings for $K$ in Appendix~\ref{ssec:hyper}) ranges and retrieves their corresponding representations $\{\mathbf{h}_k\}_{k=1}^K$ from the Segment Tree Encoder (Section~\ref{ssec:segment_tree_encoder}). The fused channel-dependent representation is:
\begin{equation}
\label{equ:h_fused}
\mathbf{h}_{\text{fused}} = \operatorname{GELU}\left(\sum_{k=1}^{K} \operatorname{Linear}^{\text{RG}}(w_k  \mathbf{h}_{\text{range-}k})\right) \in \mathbb{R}^D,
\end{equation}
where $w_k$ and $\mathbf{h}_{\text{range-}k}$ are the weight and representation of the $k$-th selected range, respectively.

\subsection{Inverse Tree Decoder}
\label{ssec:inverse_tree_decoder}

\begin{figure*}[ht]
\centering
\includegraphics[width=1\textwidth]{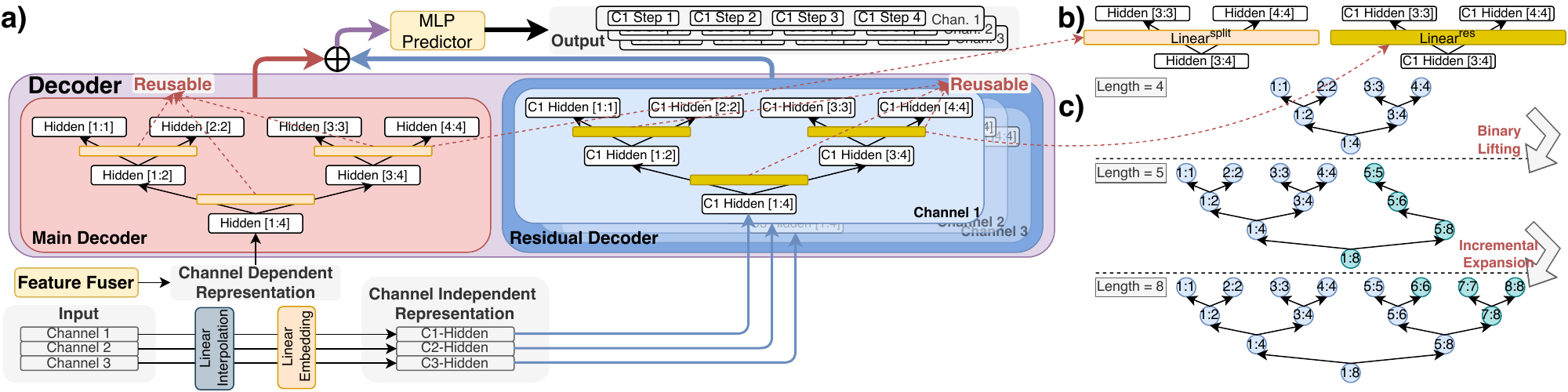}
\caption{Structure of the Inverse Tree Decoder. a) Main architecture; b) Representation decomposition method; c) Variable output length.}
\label{fig:decoder}
\end{figure*}

Mirroring the encoder's hierarchical structure, our decoder maps the compressed hidden state back to a desired forecast output length $T$ by dynamically expanding a root representation into $T$ individual time steps. The decoder adopts a dual-branch structure (Figure~\ref{fig:decoder}(a)), consisting of a channel-dependent Main Decoder and a channel-independent Residual Decoder. This design effectively balances cross-channel information with channel-specific information, an approach validated by existing studies~\cite{liu2024itransformer,han2024softs} as well as our ablation studies (Section~\ref{sec:ablation}). Unlike those fixed-size forecasting heads $ {\mathbb{R}}^{D} \Rightarrow  {\mathbb{R}}^{T}$, our Decoder decouples model parameters from $T$ by defining the transformation at the node level rather than the sequence level.

\textbf{Main Decoder: Cross-Channel Generation.}
The Main Decoder is responsible for generating the shared temporal dynamics common to all channels. Starting from the fused representation $\mathbf{z}_{\text{parent}}^{(0)} = \mathbf{h}_{\text{fused}} \in \mathbb{R}^D$ (Eq.~\ref{equ:h_fused}), it recursively splits the parent node into two child nodes at each layer:
\begin{equation}
\left[ \mathbf{z}_{\text{left}}^{(l)} ; \mathbf{z}_{\text{right}}^{(l)} \right] = \operatorname{Linear}^{\text{split}}\left( \mathbf{z}_{\text{parent}}^{(l-1)} \right) \in \mathbb{R}^{2D}.
\end{equation}
Crucially, the splitting layer $\operatorname{Linear}^{\text{split}}: \mathbb{R}^D \to \mathbb{R}^{2D}$ is shared across all layers (Figure~\ref{fig:decoder}(b)). 
By iterating this process for $H = \lceil \log_2 T \rceil$ layers, we generate $T$ leaf nodes $\{\mathbf{z}_1^{(H)}, \dots, \mathbf{z}_T^{(H)}\}$ that capture cross-channel interactions for each future time step.

\textbf{Residual Decoder: Channel-Specific Generation.}
\label{ssec:ResidualDecoder}
Complementing the Main Decoder, the Residual Decoder independently reconstructs the unique representations of each channel. It takes the channel-specific embedding $\mathbf{e}_c$ (Eq.~\ref{equ:e_c}) as its root $\mathbf{r}_{c}^{(0)}$ and applies an independent splitting process:
\begin{equation}
\left[ \mathbf{r}_{c,\text{left}}^{(l)} ; \mathbf{r}_{c,\text{right}}^{(l)} \right] = \operatorname{Linear}^{\text{res}}\left( \mathbf{r}_{c,\text{parent}}^{(l-1)} \right) \in \mathbb{R}^{2D}.
\end{equation}
This yields $T$ channel-specific leaf nodes $\{\mathbf{r}_{c,1}^{(H)}, \dots, \mathbf{r}_{c,T}^{(H)}\}$. Since this branch processes each channel in isolation, it preserves the unique distributional properties of individual channels.

\textbf{Variable-Length Output Synthesis.}
To support variable output lengths $T$, both decoders utilize a \textit{binary lifting} strategy for expansion. 
As shown in Figure~\ref{fig:decoder}(c), the tree depth $H$ is dynamically adjusted to accommodate the requested length $T$ without structural re-initialization.
The final forecast for time step $t$ in channel $c$ combines the shared and specific information:
\begin{equation}
\mathbf{o}_{c,t}=(\mathbf{z}_{t}^{(H)}+\mathbf{r}_{c,t}^{(H)}) \in \mathbb{R}^D, \, c=1 \sim C; t=1 \sim T.
\end{equation}
Here, $\mathbf{z}_{t}^{(H)}$ injects cross-channel information, while $\mathbf{r}_{c,t}^{(H)}$ provides channel-specific information. A lightweight MLP predictor then projects this state to the final scalar prediction:
\begin{equation}
\hat{y}_{c,t} = \operatorname{MLP}(\mathbf{o}_{c,t}), \quad c=1 \sim C; t=1 \sim T.
\end{equation}

\textbf{Multi-Branch Extension.}
Analogous to the Encoder, the Decoder can be generalized from a binary tree to a $B$-branch tree. This allows for shallower trees when modeling long horizons, mitigating gradient propagation issues. The branching factor $B$ is a structural hyperparameter that must be consistent across the Main and Residual branches but can differ from the Encoder's configuration. See the hyperparameter settings for $B$ in Appendix~\ref{ssec:hyper}.

\subsection{Loss Function and Model Training}
\label{ssec:training}

Standard TSF models typically adopt either Mean Absolute Error (MAE) for outlier robustness or Mean Squared Error (MSE) for high-sensitivity penalization~\cite{terven2025comprehensive}. To leverage the strengths of both, we draw on DGraFormer~\cite{yan2025dgraformer} and adopt a hybrid loss:
\begin{equation} \mathcal{L} = \alpha \cdot \mathcal{L}_{\text{MAE}} + (1 - \alpha) \cdot \mathcal{L}_{\text{MSE}}, \quad \alpha \in [0,1]. \end{equation}
Empirically, we observe $\alpha = 0.8$ yields the optimal trade-off (See Appendix~\ref{sssec:MAERate} for experimental results.). During training, we optimize the parameters by minimizing $\mathcal{L}$ over the discrete set of valid input-output configurations $\mathcal{S} = \{(L_k, T_k)\}$.

\section{Experiments}
\label{sec:exp}

We evaluate the efficacy and versatility of \method{} across two primary experimental phases: (i) \textbf{Fixed-Input Forecasting} (Section~\ref{ssec:fixed_input_forecasting}), which encompasses both standard fixed-length benchmarking (\textsf{S1}) and multi-output flexibility (\textsf{S2}). 
(ii) \textbf{Full MIOL Forecasting} (Section~\ref{ssec:full_miol_forecasting}), validating the model's native capability to handle multiple input and output lengths simultaneously (\textsf{S3}). We further conduct ablation studies to quantify each component's contribution (Section~\ref{sec:ablation}), report parameter counts, and describe representation reuse (Section~\ref{ssec:parameter_efficiency_reuse}).

\subsection{Experimental Setup}
\label{ssec:exp_setup}

\textbf{Implementation Details and Training Protocol.} 
All experiments are implemented in PyTorch 2.5 on a server with an NVIDIA RTX 5000 Ada GPU. Details in Appendix~\ref{ssec:training_details}.

\textbf{Datasets and Metrics.} 
We utilize six widely adopted benchmarks: ETTh1, ETTh2, ETTm1, ETTm2 (electricity domain), {Weather} (meteorological domain), and {Solar} (energy domain). Following standard conventions~\cite{liu2024itransformer}, we apply a 6:2:2 train-validation-test split for the ETT datasets and a 7:1:2 split for Weather and Solar. We report performance using MSE and MAE.

We compare \method{} against 11 competitive baselines from four families
\footnote{We exclude Foundation Model-based methods (e.g., LLMTime~\cite{gruver2023large} and LSTPrompt~\cite{liu2024lstprompt}) to focus on \emph{end-to-end models}. This is because a single Foundation Model's resource footprint is sufficient to deploy hundreds of the standard TSF baselines and the \method{}. Such a comparison runs counter to the core objective of this work. All baselines are detailed in Appendix~\ref{sec:detailed_descriptions}.}: (1) \textbf{Linear-based}: DLinear~\cite{zeng2023transformers}, TiDE~\cite{das2023long}, TSMixer~\cite{ekambaram2023tsmixer}, and SOFTS~\cite{han2024softs}; (2) \textbf{Transformer-based}: PatchTST~\cite{nie2023time}, iTransformer~\cite{liu2024itransformer}, DGraFormer~\cite{yan2025dgraformer}, TQNet~\cite{lin2025temporal}, and CASA~\cite{lee2025casa}; and (3) \textbf{State-Space-based}: S-Mamba~\cite{wang2025mamba}; (4) \textbf{Recurrent-Connection-based}: LSTM~\cite{hochreiter1997long}.

\subsection{Fixed-Input Forecasting (Settings \textsf{S1} and \textsf{S2})}
\label{ssec:fixed_input_forecasting}

We adhere to the standard evaluation protocol, fixing the input lookback at $L=96$ and evaluating performance across four distinct prediction horizons $T \in \{96, 192, 336, 720\}$. 
To investigate the trade-off between specialization and flexibility, we compare two training strategies:
\begin{itemize}[noitemsep, topsep=0pt, leftmargin=*]
\item \textbf{Idealized Specialist (\method{}-IS)}: We train independent \method{} models for each specific output length $T$. 
This matches the standard baseline approach and maximizes specialization for the setting \textsf{S1}.

\item \textbf{Horizon-Unified (\method{}-HU)}: 
We train a single \method{} model to jointly minimize error across all four horizons simultaneously. This validates the model's native capability to support variable outputs without retraining (supporting the setting \textsf{S2}).
\end{itemize}

\begin{table*}[ht]
\caption{Multivariate forecasting performance ($L=96$, averaged over 4 output horizons $T \in \{96, 192, 336, 720\}$).
We compare \method{}-IS and all baselines (trained specifically for each horizon $T$, following Setting \textsf{S1}) against \method{}-HU (a single unified model trained jointly to support all 4 horizons, following Setting \textsf{S2}).
\textbf{Bold} and \underline{underlined} values indicate the best and second-best results, respectively. Full results are provided in Appendix~\ref{sec:Full_FIFO_FIMO}.}
\centering
\scriptsize
\setlength{\tabcolsep}{1pt}
\resizebox{\textwidth}{!}{%
\begin{tabular}{l | >{\columncolor{green!15}}S[table-format=1.3] >{\columncolor{green!15}}S[table-format=1.3] | >{\columncolor{green!15}}S[table-format=1.3] >{\columncolor{green!15}}S[table-format=1.3] | S[table-format=1.3] S[table-format=1.3] | S[table-format=1.3] S[table-format=1.3] | S[table-format=1.3] S[table-format=1.3] | S[table-format=1.3] S[table-format=1.3] | S[table-format=1.3] S[table-format=1.3] | S[table-format=1.3] S[table-format=1.3] | S[table-format=1.3] S[table-format=1.3] | S[table-format=1.3] S[table-format=1.3] | S[table-format=1.3] S[table-format=1.3] | S[table-format=1.3] S[table-format=1.3] | S[table-format=1.3] S[table-format=1.3]}
\toprule
& \multicolumn{2}{c|}{\cellcolor{green!15}\shortstack{\method{}\\-IS (\textsf{S1})}} & \multicolumn{2}{c|}{\cellcolor{green!15}\shortstack{\method{}\\-HU (\textsf{S2})}} & \multicolumn{2}{c|}{\shortstack{DGraF.\\(2025)}} & \multicolumn{2}{c|}{\shortstack{CASA\\(2025)}} & \multicolumn{2}{c|}{\shortstack{TQNet\\(2025)}} & \multicolumn{2}{c|}{\shortstack{S-Mamba\\(2025)}} & \multicolumn{2}{c|}{\shortstack{SOFTS\\(2024)}} & \multicolumn{2}{c|}{\shortstack{iTrans.\\(2024)}} & \multicolumn{2}{c|}{\shortstack{PatchTST\\(2023)}} & \multicolumn{2}{c|}{\shortstack{TSMixer\\(2023)}} & \multicolumn{2}{c|}{\shortstack{TiDE\\(2023)}} & \multicolumn{2}{c|}{\shortstack{DLinear\\(2023)}} & \multicolumn{2}{c}{\shortstack{LSTM\\(1997)}} \\
\cmidrule(lr){2-3}\cmidrule(lr){4-5}\cmidrule(lr){6-7}\cmidrule(lr){8-9}\cmidrule(lr){10-11}\cmidrule(lr){12-13}\cmidrule(lr){14-15}\cmidrule(lr){16-17}\cmidrule(lr){18-19}\cmidrule(lr){20-21}\cmidrule(lr){22-23}\cmidrule(lr){24-25}\cmidrule(lr){26-27}
Dataset & {MSE} & {MAE} & {MSE} & {MAE} & {MSE} & {MAE} & {MSE} & {MAE} & {MSE} & {MAE} & {MSE} & {MAE} & {MSE} & {MAE} & {MSE} & {MAE} & {MSE} & {MAE} & {MSE} & {MAE} & {MSE} & {MAE} & {MSE} & {MAE} & {MSE} & {MAE}\\
\midrule
ETTm1 & \textbf{0.365} & \underline{0.379} & \underline{0.366} & \textbf{0.377} & 0.379 & 0.394 & 0.386 & 0.394 & 0.377 & 0.393 & 0.398 & 0.405 & 0.399 & 0.405 & 0.401 & 0.408 & 0.397 & 0.406 & 0.448 & 0.447 & 0.409 & 0.411 & 0.403 & 0.407 & 1.181 & 0.825 \\
\midrule
ETTm2 & \textbf{0.272} & \textbf{0.316} & \underline{0.275} & \underline{0.319} & 0.284 & 0.326 & 0.276 & \underline{0.319} & 0.277 & 0.323 & 0.288 & 0.332 & 0.287 & 0.330 & 0.288 & 0.332 & 0.287 & 0.330 & 0.289 & 0.333 & 0.358 & 0.404 & 0.350 & 0.401 & 2.562 & 1.262 \\
\midrule
ETTh1 & \textbf{0.428} & \textbf{0.425} & \textbf{0.428} & \underline{0.427} & 0.442 & 0.432 & 0.438 & 0.433 & 0.441 & 0.434 & 0.455 & 0.450 & 0.449 & 0.443 & 0.454 & 0.448 & 0.453 & 0.446 & 0.463 & 0.452 & 0.541 & 0.507 & 0.456 & 0.452 & 1.212 & 0.843 \\
\midrule
ETTh2 & \textbf{0.366} & \textbf{0.394} & \underline{0.372} & \underline{0.396} & \underline{0.372} & 0.398 & 0.373 & 0.398 & 0.378 & 0.402 & 0.381 & 0.405 & 0.373 & 0.400 & 0.383 & 0.406 & 0.385 & 0.410 & 0.402 & 0.417 & 0.611 & 0.550 & 0.559 & 0.515 & 3.283 & 1.431 \\
\midrule
Weather & \textbf{0.234} & \textbf{0.261} & \underline{0.239} & \underline{0.265} & 0.249 & 0.271 & 0.243 & 0.267 & 0.242 & 0.269 & 0.251 & 0.276 & 0.255 & 0.278 & 0.258 & 0.278 & 0.256 & 0.279 & 0.256 & 0.278 & 0.270 & 0.320 & 0.265 & 0.317 & 0.406 & 0.401 \\
\midrule
Solar & \underline{0.204} & \textbf{0.239} & 0.207 & 0.242 & 0.216 & \textbf{0.239} & 0.222 & 0.244 & \textbf{0.198} & 0.256 & 0.240 & 0.272 & 0.229 & 0.256 & 0.233 & 0.262 & 0.236 & 0.266 & 0.261 & 0.297 & 0.347 & 0.417 & 0.330 & 0.401 & 0.498 & 0.436 \\
\bottomrule
\end{tabular}%
}
\label{tab:fixdynamic_avg}
\end{table*}

As shown in Table~\ref{tab:fixdynamic_avg}, the horizon-specific \method{}-IS establishes a new SOTA. Crucially, the horizon-unified \method{}-HU incurs negligible performance loss (and occasionally exceeds \method{}-IS), despite optimizing for multiple output lengths simultaneously.
Although LSTM~\cite{hochreiter1997long} supports arbitrary lengths, their performance significantly lags behind other modern models.
This resilience confirms that our Inverse Tree Decoder effectively decouples model parameters from output length, enabling flexible deployment without substantial accuracy degradation typical of heuristic adaptations.

\subsection{Full MIOL Forecasting (Setting \textsf{S3})}
\label{ssec:full_miol_forecasting}

To validate the model's adaptability in dynamic deployment environments, we evaluate performance under the Full MIOL setting, where both input lengths $L$ and output lengths $T$ vary during inference.

We compare our natively unified \method{}-MIOL against two categories of baselines:
(1) \method{}-IS: a separate model is trained specifically for every distinct $(L, T)$ pair.
(2) {Heuristic Adapters}: Traditional strategies that force fixed-size models to handle variable lengths without training multiple times.
For input adaptation, we apply \emph{Trimming} or \emph{Padding}.
For output adaptation, we employ \emph{Autoregression} or \emph{Trimming}.
This yields four combinatorial heuristics: \textsc{Trim}-\textsc{Reg}, \textsc{Trim}-\textsc{Trim}, \textsc{Pad}-\textsc{Reg}, and \textsc{Pad}-\textsc{Trim}.
We provide more details in Appendix~\ref{sec:Full_MIOL}.

Table~\ref{tab:multilen_avg} details the trade-offs across accuracy (MSE/MAE), inference latency, and memory usage.
Remarkably, the unified {\method{}-MIOL} matches the accuracy of the specialized \method{}-IS, effectively eliminating the need for multi-model maintenance without substantial performance degradation. 
In contrast, heuristics incur severe penalties. The input-side \textsc{Trim}-$\star$ methods achieve low latency but incur significant accuracy loss due to discarding the first half of the input. Meanwhile, the output-side $\star$-\textsc{Reg} methods reduce memory usage at the cost of accuracy and higher latency (from iterative forecasting). Among heuristics, \textsc{Pad}-\textsc{Trim} achieves the best forecasting accuracy by avoiding both input truncation and error accumulation. However, it incurs substantially higher latency and memory usage due to modeling at the maximum length, whereas {\method{}-MIOL} incurs no such costs. Applying these heuristics to previous SOTA models yields consistent conclusions.\footnote{As detailed in Appendix~\ref{sec:MIOL_Other}, applying these heuristics to SOTA models (e.g., CASA, DGraFormer) results in even sharper performance drops, confirming that \method{}'s resilience stems from its native architectural elasticity (Segment Tree + Inverse Tree) rather than simple data augmentation.}

\begin{table*}[ht]
\caption{Performance trade-offs in the {Full MIOL} setting (averaged over $L \in \{48, 96\}$ and $T \in \{48, 96\}$). We compare the unified {\method{}-MIOL} against the idealized specialist \textcolor{gray}{\method{}-IS} (trained for each $(L, T)$ pair, as the baseline) and four heuristic adapters. \emph{Time} and \emph{MEM} denote inference latency (s) and memory usage (GB). Full results and the results across 8 pairs from $L = \{48, 96, 144\}$ and $T = \{96, 192, 336, 720\}$ are provided in Appendix~\ref{ssec:full_MIOL_res}.}
\centering
\scriptsize
\setlength{\tabcolsep}{1pt}
\renewcommand{\arraystretch}{0.95}
\resizebox{\textwidth}{!}{%
\begin{tabular}{l | >{\color{gray}}S[table-format=1.3] >{\color{gray}}S[table-format=1.3] >{\color{gray}}S[table-format=1.3] >{\color{gray}}S[table-format=2.2] | >{\columncolor{green!15}}S[table-format=1.3] >{\columncolor{green!15}}S[table-format=1.3] >{\columncolor{green!15}}S[table-format=1.3] >{\columncolor{green!15}}S[table-format=2.2] | S[table-format=1.3] S[table-format=1.3] S[table-format=1.3] S[table-format=2.2] | S[table-format=1.3] S[table-format=1.3] S[table-format=1.3] S[table-format=2.2] | S[table-format=1.3] S[table-format=1.3] S[table-format=1.3] S[table-format=2.2] | S[table-format=1.3] S[table-format=1.3] S[table-format=1.3] S[table-format=2.2]}
\toprule
&
\multicolumn{4}{c|}{\textcolor{gray}{\shortstack{\method{}\\-IS (\textsf{S1})}}} &
\multicolumn{4}{c|}{\cellcolor{green!15}\shortstack{\method{}\\-MIOL (\textsf{S3})}} &
\multicolumn{4}{c|}{\shortstack{\method{}\\+\textsc{Trim}-\textsc{Reg}}} &
\multicolumn{4}{c|}{\shortstack{\method{}\\+\textsc{Trim}-\textsc{Trim}}} &
\multicolumn{4}{c|}{\shortstack{\method{}\\+\textsc{Pad}-\textsc{Reg}}} &
\multicolumn{4}{c}{\shortstack{\method{}\\+\textsc{Pad}-\textsc{Trim}}} \\
\cmidrule(lr){2-5}\cmidrule(lr){6-9}\cmidrule(lr){10-13}\cmidrule(lr){14-17}\cmidrule(lr){18-21}\cmidrule(lr){22-25}
Dataset & {\textcolor{gray}{MSE}} & {\textcolor{gray}{MAE}} & {\textcolor{gray}{Time}} & {\textcolor{gray}{MEM}} & {\cellcolor{green!15}MSE} & {\cellcolor{green!15}MAE} & {\cellcolor{green!15}Time} & {\cellcolor{green!15}MEM} & {MSE} & {MAE} & {Time} & {MEM} & {MSE} & {MAE} & {Time} & {MEM} & {MSE} & {MAE} & {Time} & {MEM} & {MSE} & {MAE} & {Time} & {MEM} \\
\midrule
ETTh1 & 0.353 & 0.374 & 0.248 & 9.71 & \textbf{0.356} & \textbf{0.375} & 0.244 & 9.71 & 0.370 & \underline{0.383} & 0.210 & 6.48 & 0.362 & \underline{0.383} & 0.121 & 12.74 & 0.370 & 0.389 & 0.633 & 6.69 & \underline{0.359} & 0.386 & 0.465 & 12.91 \\
\midrule
ETTh2 & 0.256 & 0.314 & 0.261 & 9.68 & \textbf{0.258} & \textbf{0.315} & 0.257 & 9.66 & 0.268 & 0.322 & 0.214 & 6.45 & 0.265 & \underline{0.316} & 0.113 & 12.65 & 0.263 & 0.331 & 0.628 & 6.90 & \underline{0.259} & 0.319 & 0.414 & 12.95 \\
\midrule
ETTm1 & 0.337 & 0.354 & 0.262 & 9.67 & \textbf{0.343} & \textbf{0.357} & 0.270 & 9.67 & 0.446 & 0.412 & 0.231 & 6.49 & 0.403 & 0.389 & 0.110 & 12.64 & 0.407 & 0.397 & 0.720 & 6.72 & \underline{0.370} & \underline{0.374} & 0.403 & 12.91 \\
\midrule
ETTm2 & 0.153 & 0.237 & 0.251 & 9.76 & \textbf{0.155} & \textbf{0.237} & 0.261 & 9.72 & \underline{0.167} & \underline{0.256} & 0.221 & 6.47 & 0.169 & 0.256 & 0.125 & 12.68 & 0.169 & 0.262 & 0.695 & 6.77 & \underline{0.167} & 0.259 & 0.389 & 12.91 \\
\midrule
Weather & 0.145 & 0.182 & 0.258 & 13.95 & \underline{0.147} & \textbf{0.184} & 0.252 & 13.97 & 0.192 & 0.229 & 0.242 & 9.39 & 0.168 & 0.205 & 0.116 & 18.48 & \underline{0.147} & 0.189 & 0.672 & 9.48 & \textbf{0.146} & \textbf{0.184} & 0.400 & 18.64 \\
\midrule
Solar & 0.243 & 0.286 & 0.265 & 16.83 & \textbf{0.240} & \underline{0.285} & 0.268 & 16.91 & 0.602 & 0.586 & 0.257 & 11.34 & 0.302 & 0.336 & 0.122 & 22.45 & \underline{0.244} & 0.291 & 0.685 & 11.39 & \underline{0.244} & \textbf{0.281} & 0.412 & 22.49 \\
\bottomrule
\end{tabular}%
}
\label{tab:multilen_avg}
\end{table*}

\subsection{Ablation Study} 
\label{sec:ablation}

\begin{figure}[ht]
\centering
\includegraphics[width=1\textwidth]{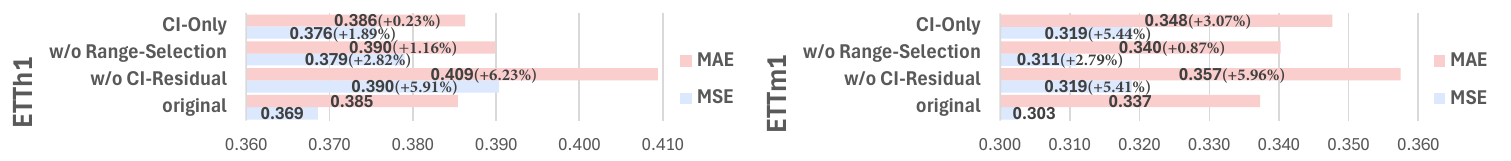}
\caption{Ablation results with $L=96, T=96$ (full results and analysis in Appendix~\ref{ssec:ablation_fig}).}
\label{fig:AblationRes_small}
\end{figure}

To quantify the contribution of individual components, we evaluate 3 variants against the original model (figures in Appendix~\ref{ssec:ablation_fig}):
(1) \textsf{w/o CI-Residual}: Removes the channel-independent residual branch. This tests whether explicitly modeling shared global trends complements the channel-specific tree representations.
(2) \textsf{w/o Range-Selection}: Bypasses the Range Weighter and Feature Fuser, utilizing only the Segment Tree's root node. This verifies the necessity of our multi-scale aggregation compared to using a single global representation.
(3) \textsf{CI-Only}: Retains only the channel-independent residual structure. This baseline quantifies the performance gain attributed to our tree-based channel-dependent modeling. Figure~\ref{fig:AblationRes_small} shows the full \method{} yields the lowest error.


\subsection{Parameter Efficiency and Representation Reuse}
\label{ssec:parameter_efficiency_reuse}

\textbf{Parameter efficiency.}
To illustrate the parameter-sharing benefit of \method{}'s native MIOL support, we select four representative baselines: the SOTA models CASA (Transformer-based) and SOFTS (Linear-based), and the classic models PatchTST (Transformer-based) and DLinear (Linear-based). These baselines do not natively support MIOL, and their official implementations use the standard input length $L=96$ and provide no configurations for other input lengths. We therefore compare them with \method{}-HU, which shares one model across multiple output lengths at the same input length. 

\begin{figure}[ht]
\centering
\includegraphics[width=0.8\textwidth]{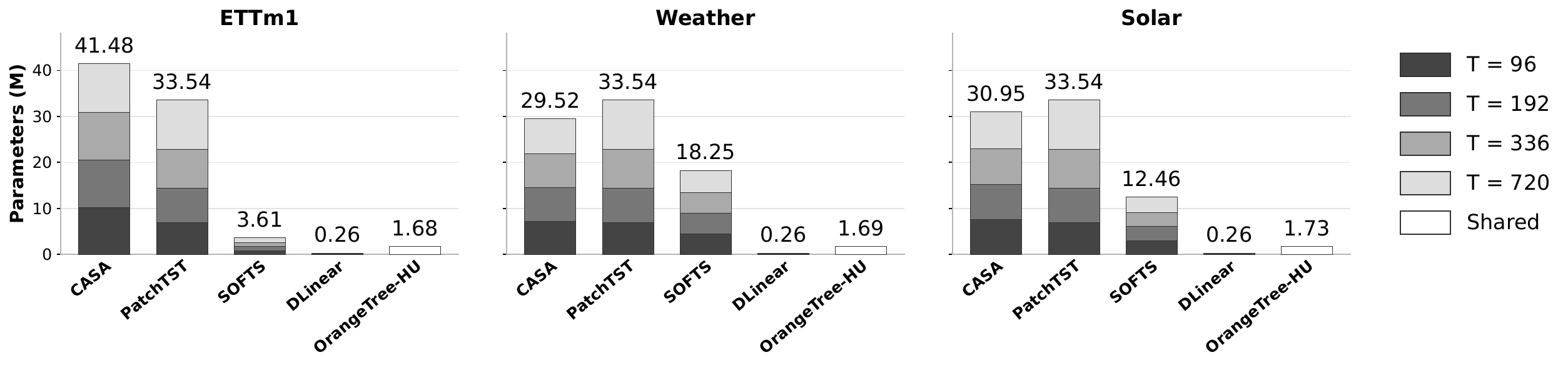}
\caption{Parameter counts on ETTm1, Weather, and Solar using official baseline hyperparameters with input length $L=96$ and output lengths $T \in \{96,192,336,720\}$.}
\label{fig:parameter_counts}
\end{figure}

As illustrated in Figure~\ref{fig:parameter_counts}, we compare both single-model parameter counts and the total required to cover all output lengths. For a single model, \method{}-HU has fewer parameters than CASA and PatchTST, while its comparison with SOFTS varies across datasets. When deploying models for all output lengths, \method{}-HU requires far fewer total parameters than CASA, PatchTST, and SOFTS, while also achieving better average MSE and MAE (Table~\ref{tab:fixdynamic_avg}). These results show that \method{}-HU offers both better forecasting performance and an advantage in parameter count over these strong baselines. Although DLinear, an extremely lightweight classic Linear model, has fewer total parameters than \method{}-HU, it lags far behind \method{} and other SOTA models in forecasting performance.

\begin{figure}[ht]
\centering
\includegraphics[width=0.7\textwidth]{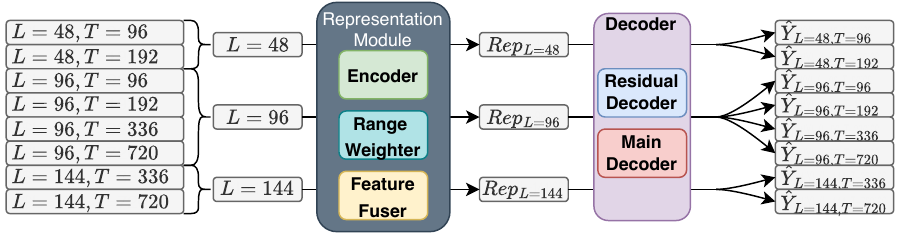}
\caption{Representation reuse for different input lengths $L$ and output lengths $T$ in one \method{} model.}
\label{fig:representation_reuse}
\end{figure}

\textbf{Representation reuse.}
\method{}'s native MIOL support not only enables parameter sharing but also allows representation reuse during inference. Length-fixed baselines train a separate model for each $(L,T)$ pair, so they cannot share one model across output lengths even when the input length is the same. As illustrated in Figure~\ref{fig:representation_reuse}, a single \method{} model covers multiple trained input-output length pairs $(L,T)$ without switching models, and the representation of each input is shared across its output lengths. For the same input of length $L$, the Representation Module runs once. The Decoder reuses the representation to generate forecasts for each output length $T$.

\section{Conclusion}
\label{sec:conclusion}

We introduce \method{}, a linear and tree-based architecture decoupling model parameters from sequence length. \method{} achieves state-of-the-art results on fixed-horizon benchmarks and natively resolves the {Multiple Input and Output Length} (MIOL) challenge, without parameter inflation, computational redundancy, or severe performance degradation. \footnote{Limitation and future work are discussed in Appendix~\ref{sec:limitation}.}


\clearpage
\newpage

\section*{Acknowledgments}
This work was supported by the National Natural Science Foundation of China (No. 62402420), Fundamental Research Funds for Central Universities (Grant No. 226202400109), and Beijing Natural Science Foundation (No. L233034). Huan Li is the corresponding author of this paper.

\bibliographystyle{unsrt}
\bibliography{ref}

\newpage
\appendix

\section{Impact Statement}
This work contributes to sustainable and accessible AI. By proving that a linear architecture can rival heavy Transformers, \method{} drastically reduces the computational and storage footprint required for high-performance forecasting. This efficiency facilitates deployment on resource-constrained devices, directly supporting Green AI goals by lowering energy consumption. We do not foresee any negative societal impacts arising from this research.

\section{Ethics Statement}
We have ensured that all datasets used in the experiments are publicly available, and we have carefully considered the ethical implications of our work. Therefore, we believe that our research does not present any ethical or moral risks.

\section{Detailed Descriptions of All Baselines}
\label{sec:detailed_descriptions}
\subsection{Linear Models}
\begin{itemize}[leftmargin=*]
\item DLinear~\cite{zeng2023transformers}: It follows the paradigm of seasonal-trend decomposition. It employs two linear layers to predict the seasonal and trend parts separately, which are then summed to produce the final forecast.
\item TiDE~\cite{das2023long}: It utilizes a Time-series Dense Encoder architecture. TiDE encodes historical time series across all channels using an MLP and then decodes them into future values with another MLP.
\item TSMixer~\cite{ekambaram2023tsmixer}: Through hierarchical patch alignment and cross-channel online calibration, it enhances long-sequence and multivariate modeling capabilities, and maintains low computational cost.
\item SOFTS~\cite{han2024softs}: Through the STAR module with an ``aggregate-distribute-fuse'' mechanism, it achieves channel-independent and channel-dependent modeling, thereby improving performance in multivariate TSF.
\end{itemize}

\subsection{Transformer Models}
\begin{itemize}[leftmargin=*]
\item PatchTST~\cite{nie2023time}: It uses a patching mechanism to reduce computational and storage costs, enabling the same window length to capture longer TS dependencies. PatchTST adopts a channel-independent architecture to reduce cross-channel interference. Simultaneously, it also designs a masked reconstruction task to assist the model in learning temporal representations.
\item iTransformer~\cite{liu2024itransformer}: It inverts the operating dimensions of the Transformer's attention and FFN, embedding each channel as a token, thereby using attention to explicitly model cross-channel correlations, and using the FFN to learn intra-channel representations.
\item DGraFormer~\cite{yan2025dgraformer}: It employs dynamic correlation-aware graph learning to capture correlations across different time windows, and simultaneously uses a multi-scale temporal transformer to extract temporal patterns from sequences of different scales.
\item TQNet~\cite{lin2025temporal}: This technique employs periodically shifted learnable vectors as the Q in the attention mechanism to capture global cross-channel patterns, while generating K and V from the original input data to encode local correlations.
\item CASA~\cite{lee2025casa}: It adopts a multi-layer CNN-based autoencoder structure, aiming to replace the computation of K and Q in the traditional attention structure, thereby improving prediction performance and reducing complexity.
\end{itemize}

\subsection{State-Space Models}
\begin{itemize}[leftmargin=*]
\item S-Mamba~\cite{wang2025mamba}: It first embeds each timestep of each channel through a linear layer, then utilizes a bidirectional Mamba layer to extract correlations between channels. Finally, it generates the prediction results through a linear layer.
\end{itemize}

\subsection{Recurrent-Connection Models}
\begin{itemize}[leftmargin=*]
\item LSTM~\cite{hochreiter1997long}: It is an improved variant of RNN~\cite{elman1990finding} that introduces a memory cell and gating mechanisms to learn both short-term and long-term dependencies. It natively supports arbitrary input and output lengths via its recurrent connections.
\end{itemize}

\section{Model Details}
\label{sec:model_details}
\subsection{Computation Details}
\label{ssec:computation_details}
\subsubsection{Hierarchical Representation Extraction in the Encoder}
\label{sssec:RepExt}
Regardless of which range's representation within the historical series is to be extracted, the process ultimately requires aggregation to the root node, as shown in Figure~\ref{fig:encoder_agg}. Several examples are illustrated below:
\begin{itemize}[leftmargin=*]
\item Getting the representation of the range $[1:4]$: The representation for range $[1:4]$ has been pre-computed, and since $[1:4]$ is the root node, we use it directly, as shown in Figure~\ref{fig:encoder_agg} (a).

\item Getting the representation of the range $[2:4]$: We need node $[2:2]$ and node $[3:4]$ to compose this range. However, node $[2:2]$ and node $[3:4]$ cannot be aggregated directly because they belong to different layers. To bring node $[2:2]$ and node $[3:4]$ to the same layer, we need to propagate the representation of node $[2:2]$ to its parent node and pad the nodes that participate in the computation but are not selected with zeros, as shown in Figure~\ref{fig:encoder_agg} (b). The formula is:
\begin{equation}
{\mathbf{h}}_{\left\lbrack  2 : 4\right\rbrack  } = {\operatorname{Linear}}^{\mathrm{{ag}}}\left( \left\lbrack  {{\operatorname{Linear}}^{\mathrm{{ag}}}\left( \left\lbrack  {{\mathbf{h}}_{\text{ zeros }};{\mathbf{h}}_{\left\lbrack  2 : 2\right\rbrack  }}\right\rbrack  \right) ;{\mathbf{h}}_{\left\lbrack  3 : 4\right\rbrack  }}\right\rbrack  \right)  \in  {\mathbb{R}}^{D}\text{ . }
\end{equation}

\item Getting the representation of the range $[2:3]$: Similarly, as shown in Figure~\ref{fig:encoder_agg} (c), the formula is:
\begin{equation}
{\mathbf{h}}_{\left\lbrack  2 : 3\right\rbrack  } = {\operatorname{Linear}}^{\mathrm{{ag}}}\left( \left\lbrack  {{\operatorname{Linear}}^{\mathrm{{ag}}}\left( \left\lbrack  {{\mathbf{h}}_{\text{ zeros }};{\mathbf{h}}_{\left\lbrack  2 : 2\right\rbrack  }}\right\rbrack  \right) ;{\operatorname{Linear}}^{\mathrm{{ag}}}\left( \left\lbrack  {{\mathbf{h}}_{\left\lbrack  3 : 3\right\rbrack  };{\mathbf{h}}_{\text{ zeros }}}\right\rbrack  \right) }\right\rbrack  \right)  \in  {\mathbb{R}}^{D}\text{ . }
\end{equation}
\end{itemize}

\begin{figure}[ht]
\centering\includegraphics[width=0.95\textwidth]{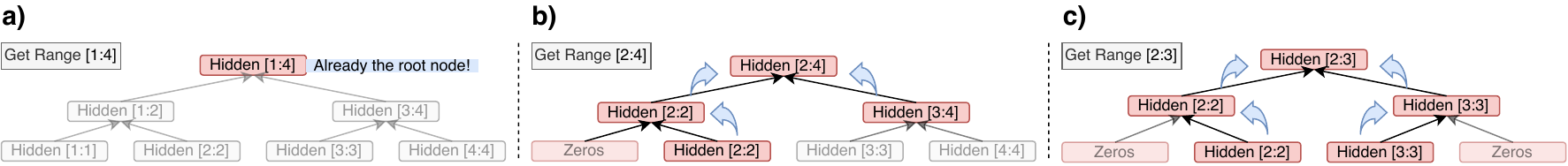}
\caption{Example: Encoder Aggregation Process.}
\label{fig:encoder_agg}
\end{figure}

\subsubsection{Weight Calculation in the Range Weighter}
\label{sssec:WeightCalc}
\begin{figure}[ht]
\centering\includegraphics[width=0.5\textwidth]{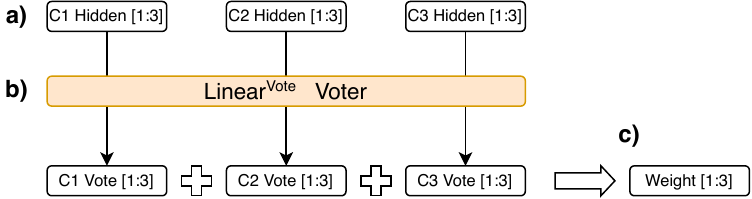}
\caption{Weight calculation in the Range Weighter (Taking range [1:3] as an example).}
\label{fig:weighter_vote}
\end{figure}

When calculating the weight for a specific range, first, the hidden states of that range across all channels are retrieved, as shown in Figure~\ref{fig:weighter_vote} (a). Then, this hidden state is projected through a shared linear layer to obtain the vote (a scalar) of each channel for that range, as shown in Figure~\ref{fig:weighter_vote} (b). Finally, the votes from all channels for that range are summed to obtain the final weight of the range, as shown in Figure~\ref{fig:weighter_vote} (c).

\subsection{Hyperparameters}
\label{ssec:hyper}
\subsubsection{The top-K ranges of Feature Fuser}
Adjusting top-$K$ typically does not directly fix a specific $K$. This is because, in the case of MIOL, the number of ranges required for representation differs between longer and shorter inputs. Therefore, we use a $K\text{-Rate}$ to calculate the $K$ for a given input length. For an input of length $L$, the total number of ranges is:
\begin{equation}
{N}_{range} = \frac{\left( {L + 1}\right)  \times  L}{2}.
\end{equation}
The parameter $K$ in top-$K$ is:
\begin{equation}
K = {N}_{range} \times  K\text{-Rate}.
\end{equation}

\subsubsection{Branch Number of Encoder and Decoder}
For convenience, we usually synchronize the adjustment of the branch number ($B$) in both the Encoder and Decoder. A larger $B$ increases the parameter size, which allows the model to capture details. Conversely, a smaller $B$ reduces parameter size and can enhance model robustness. Therefore, selecting an appropriate $B$ during training is essential.

\subsubsection{Hyperparameter Settings}
\label{sssec:hyperparameter_settings}

Table~\ref{tab:core_hyperparameters} summarizes the core hyperparameters of \method{} and their search spaces or fixed values. We tune $B$ and $K$ separately for each dataset and input-output length pair $(L,T)$. We search the branch number $B$ over $\{2,3,4,5\}$. For top-$K$ selection, we search either $K\text{-Rate}$ over $\{0.1,0.01,0.001\}$ or $K$ over $\{4,16,64\}$. The hidden size $D$ and the Range Weighter's hidden size $D'$ are fixed at 384 and 32, respectively. The hybrid loss weight is set to $\alpha=0.8$, assigning weights of 0.8 to MAE and 0.2 to MSE.

\begin{table}[ht]
\caption{Core hyperparameters of \method{}.}
\label{tab:core_hyperparameters}
\centering
\small
\renewcommand{\arraystretch}{1.2}
\setlength{\aboverulesep}{1.5pt}
\setlength{\belowrulesep}{1.5pt}
\renewcommand{\tabularxcolumn}[1]{m{#1}}
\begin{tabularx}{\textwidth}{@{}>{\centering\arraybackslash}m{0.18\textwidth}>{\raggedright\arraybackslash}m{0.34\textwidth}>{\centering\arraybackslash}X@{}}
\toprule
\textbf{Hyperparameter} & \textbf{Meaning} & \textbf{Value} \\
\midrule
$B$ & Branch Number & Search in $\{2,3,4,5\}$ \\
\midrule
$K$ & Top-$K$ Selection & \shortstack[l]{Search $K\text{-Rate}$ in $\{0.1,0.01,0.001\}$\\or $K$ in $\{4,16,64\}$} \\
\midrule
$D$ & Hidden Size & 384 \\
\midrule
$D'$ & Hidden Size for the Range Weighter & 32 \\
\midrule
$\alpha$ & $\mathcal{L}=\alpha\cdot\mathrm{MAE}+(1-\alpha)\cdot\mathrm{MSE}$ & 0.8 \\
\bottomrule
\end{tabularx}
\end{table}

\subsubsection{Hybrid Loss}
\label{sssec:MAERate}

The function of the hybrid loss is:
\begin{equation} \mathcal{L} = \alpha \cdot \mathcal{L}_{\text{MAE}} + (1 - \alpha) \cdot \mathcal{L}_{\text{MSE}}, \quad \alpha \in [0,1]. \end{equation}
As shown in Table~\ref{tab:mae_rate}, setting $\alpha=0.8$ consistently yields the most stable and optimized MSE+MAE improvements across datasets.

\begin{table*}[ht]
\caption{Effect of $\alpha$ on model performance ($L=96, T=96$).}
\centering
\scriptsize
\setlength{\tabcolsep}{3pt}
\resizebox{\textwidth}{!}{%
\begin{tabular}{l | S[table-format=1.3] S[table-format=1.3] S[table-format=1.3] | S[table-format=1.3] S[table-format=1.3] S[table-format=1.3] | S[table-format=1.3] S[table-format=1.3] S[table-format=1.3] | S[table-format=1.3] S[table-format=1.3] S[table-format=1.3] | S[table-format=1.3] S[table-format=1.3] S[table-format=1.3] | S[table-format=1.3] S[table-format=1.3] S[table-format=1.3]}
\toprule
Dataset & \multicolumn{3}{c|}{$\alpha=0.0$} & \multicolumn{3}{c|}{$\alpha=0.2$} & \multicolumn{3}{c|}{$\alpha=0.4$} & \multicolumn{3}{c|}{$\alpha=0.6$} & \multicolumn{3}{c|}{$\alpha=0.8$} & \multicolumn{3}{c}{$\alpha=1.0$} \\
\cmidrule(lr){2-4}\cmidrule(lr){5-7}\cmidrule(lr){8-10}\cmidrule(lr){11-13}\cmidrule(lr){14-16}\cmidrule(lr){17-19}
 & {MSE} & {MAE} & {MSE+MAE} & {MSE} & {MAE} & {MSE+MAE} & {MSE} & {MAE} & {MSE+MAE} & {MSE} & {MAE} & {MSE+MAE} & {MSE} & {MAE} & {MSE+MAE} & {MSE} & {MAE} & {MSE+MAE} \\
\midrule
ETTh1 & 0.376 & 0.395 & 0.771 & 0.372 & 0.392 & 0.764 & 0.368 & 0.388 & 0.756 & 0.370 & 0.388 & 0.758 & 0.369 & 0.385 & \textbf{0.754} & 0.382 & 0.389 & 0.771 \\
\midrule
ETTm1 & 0.301 & 0.350 & 0.651 & 0.299 & 0.344 & 0.643 & 0.299 & 0.341 & \textbf{0.640} & 0.302 & 0.340 & 0.642 & 0.303 & 0.337 & \textbf{0.640} & 0.311 & 0.337 & 0.648 \\
\midrule
Weather & 0.158 & 0.219 & 0.377 & 0.157 & 0.205 & 0.362 & 0.154 & 0.202 & 0.356 & 0.156 & 0.198 & 0.354 & 0.149 & 0.187 & \textbf{0.336} & 0.154 & 0.196 & 0.350 \\
\bottomrule
\end{tabular}%
}
\label{tab:mae_rate}
\end{table*}

\subsection{Ablation}
\label{ssec:ablation_fig}
\begin{figure}[ht]
\centering
\includegraphics[width=1\textwidth]{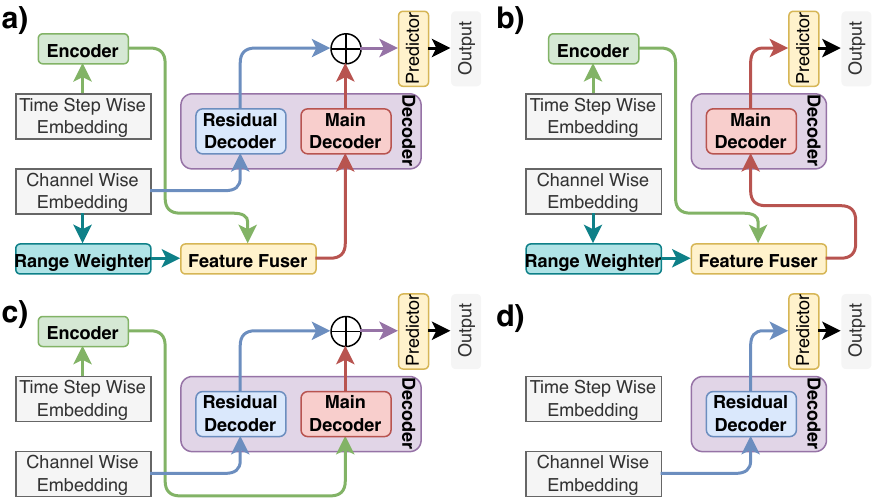}
\caption{Ablation study: a) original structure; b) w/o CI-Residual; c) w/o Range-Selection; and d) CI-Only.}
\label{fig:ablation}
\end{figure}

To validate the contribution of each component in \method{}, we design three architectural variants as illustrated in Figure~\ref{fig:ablation}. Each variant systematically removes or isolates specific components to quantify their impact on forecasting accuracy.

\begin{figure}[ht]
\centering
\includegraphics[width=1\textwidth]{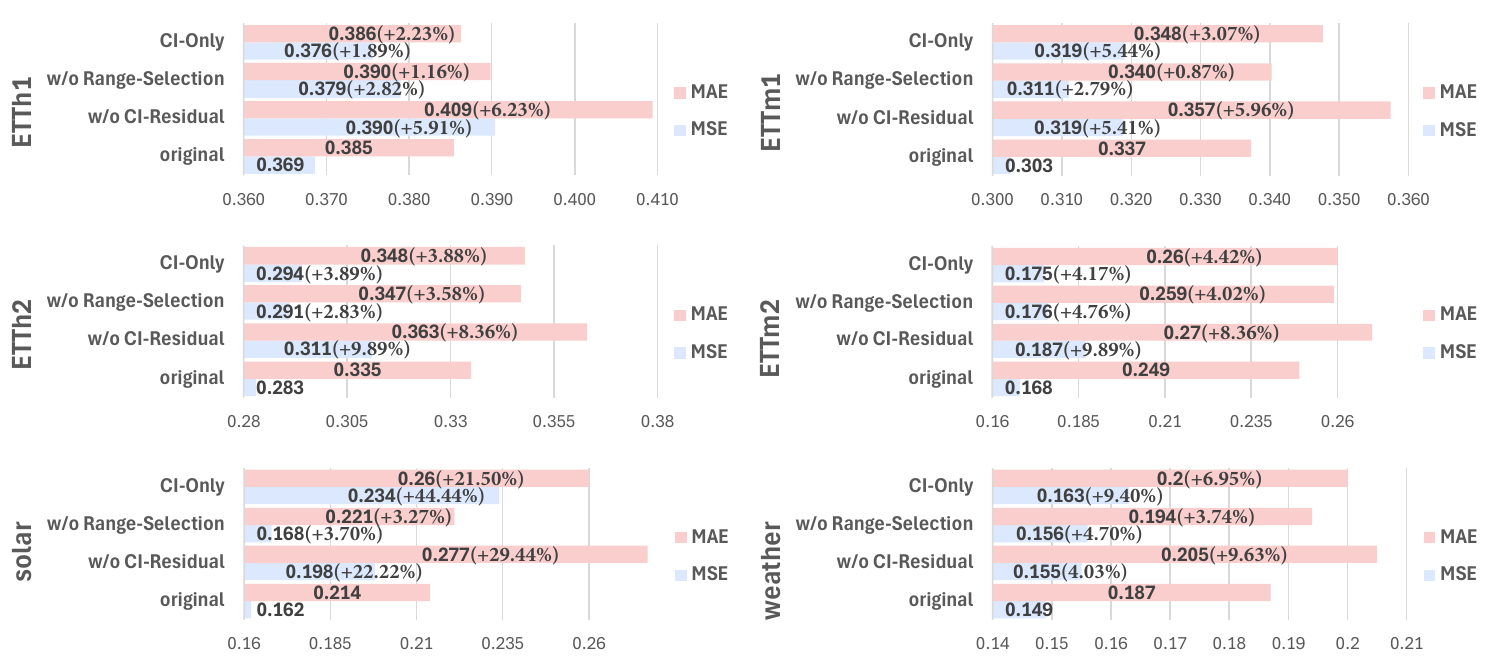}
\caption{Ablation results ($L=96$, $T=96$).}
\label{fig:AblationRes}
\end{figure}
 
As shown in Figure~\ref{fig:AblationRes}, 
the full \method{} consistently yields the lowest error. The performance degradation in \textsf{w/o Range-Selection} confirms that global representation alone is insufficient, validating our top-$K$ selection mechanism for capturing localized temporal shifts. Furthermore, the significant drop in \textsf{w/o CI-Residual} underscores the importance of the dual-branch design, proving that balancing shared channel dynamics with independent residual correction is critical for robust forecasting.


\subsection{Training Details}
\label{ssec:training_details}
\subsubsection{Hardware and Software}
All experiments are conducted on a consistent computing environment to ensure reproducibility. Table~\ref{tab:hardware_software} summarizes the complete hardware and software specifications.

\begin{table}[ht]
\centering
\caption{Hardware and software configuration.}
\renewcommand{\arraystretch}{0.8}
\begin{tabular}{rl}
\toprule
\textbf{Component} & \textbf{Specification} \\
\midrule
CPU & Intel(R) Xeon(R) w9-3495X \\
GPU & NVIDIA RTX 5000 Ada (32GB, Ada Lovelace architecture) \\
NVIDIA Driver & Version 570.195.03 \\
CUDA & Version 12.8 \\
OS & Ubuntu 24.04.3 LTS \\
Deep Learning Framework & PyTorch 2.5.1+cu121 \\
\bottomrule
\end{tabular}
\label{tab:hardware_software}
\end{table}

\subsubsection{Variable-Length Training}

\begin{figure}[ht]
\centering\includegraphics[width=0.9\textwidth]{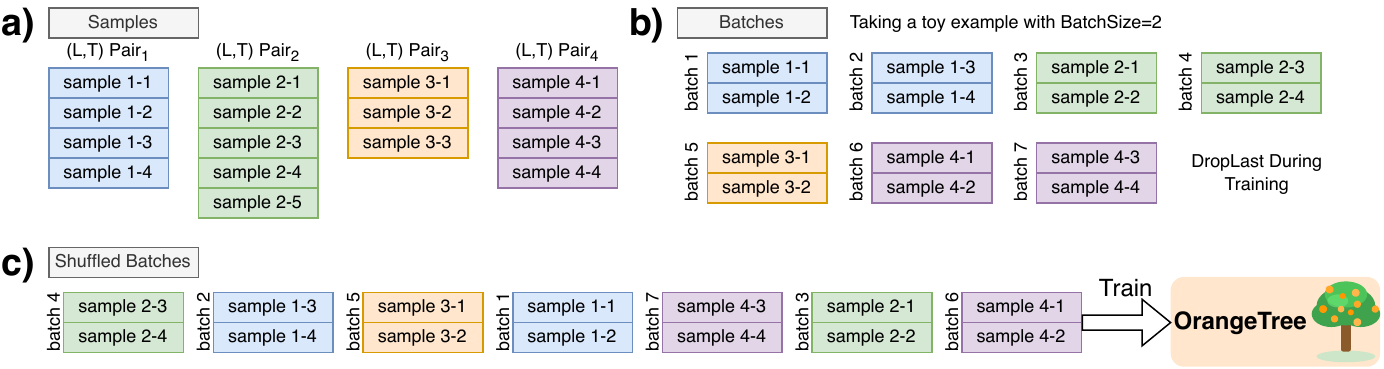}
\caption{A toy example of variable-length training.}
\label{fig:train_detail}
\end{figure}

For scenarios involving variable lengths (Settings \textsf{S2} and \textsf{S3}), we adopt a training strategy where sequence lengths are uniform within each batch but shuffled across batches. Specifically, we define different input-output length pairs as $(L, T)~\text{Pair}_n$, where each $(L, T)~\text{Pair}$ contains a slightly varying number of samples, as shown in Figure~\ref{fig:train_detail}(a). Subsequently, we create independent batches for each $(L, T)~\text{Pair}$ (with $\text{DropLast}=\text{True}$ during training and $\text{DropLast}=\text{False}$ during validation and testing), as shown in Figure~\ref{fig:train_detail}(b). Finally, we shuffle all batches and feed them into \method{} for training, as shown in Figure~\ref{fig:train_detail}(c).

\label{sssec:VarLenTraining}

\section{Limitation and Future Work}
\label{sec:limitation}
Although \method{} theoretically supports arbitrary horizons, we observe a practical ``Generalist-Specialist Gap'' between fully-arbitrary-length and length-specialized training. 

As shown in Table~\ref{tab:arbitrary_length}, the performance drop of MIOL (\textsf{S3}) compared to IS (\textsf{S1}) is negligible. However, under \textsf{Arbitrary Length} training, a significant performance degradation appears, though it is not entirely out of control. This is the so-called "Generalist-Specialist Gap". We highlight this gap as a strategic avenue for achieving extremely elastic temporal context windows in future time series forecasting models.

\begin{table}[ht]
\caption{Performance comparison between MIOL and Arbitrary Length. IS (\textsf{S1}) trains one model per length as the baseline. MIOL (\textsf{S3}) is trained only on $L, T \in \{48, 96\}$. Arbitrary Length trains on continuous lengths $L, T \in [48, 96]$.}
\centering
\renewcommand{\arraystretch}{0.7}
\begin{tabular}{l l | >{\color{gray}}S[table-format=1.3] >{\color{gray}}S[table-format=1.3] | S[table-format=1.3] S[table-format=1.3] | S[table-format=1.3] S[table-format=1.3]}
\toprule
\multirow{2}{*}{Dataset} & \multirow{2}{*}{L-T} & \multicolumn{2}{c|}{\textcolor{gray}{IS (S1)}} & \multicolumn{2}{c|}{MIOL (S3)} & \multicolumn{2}{c}{Arbitrary Length} \\
\cmidrule(lr){3-4} \cmidrule(lr){5-6} \cmidrule(lr){7-8}
 & & {MSE} & {MAE} & {MSE} & {MAE} & {MSE} & {MAE} \\
\midrule
ETTh1 & 48-48 & 0.338 & 0.362 & 0.339 & 0.363 & 0.381 & 0.408 \\
 & 48-96 & 0.376 & 0.389 & 0.381 & 0.390 & 0.431 & 0.439 \\
 & 96-48 & 0.328 & 0.361 & 0.331 & 0.361 & 0.372 & 0.390 \\
 & 96-96 & 0.369 & 0.385 & 0.371 & 0.387 & 0.421 & 0.418 \\
 & AVG & 0.353 & 0.374 & 0.356 & 0.375 & 0.401 & 0.414 \\
\midrule
ETTm1 & 48-48 & 0.392 & 0.383 & 0.412 & 0.388 & 0.580 & 0.481 \\
 & 48-96 & 0.382 & 0.380 & 0.379 & 0.380 & 0.519 & 0.469 \\
 & 96-48 & 0.269 & 0.314 & 0.273 & 0.317 & 0.324 & 0.357 \\
 & 96-96 & 0.303 & 0.337 & 0.306 & 0.341 & 0.356 & 0.379 \\
 & AVG & 0.337 & 0.354 & 0.343 & 0.357 & 0.445 & 0.422 \\
\bottomrule
\end{tabular}
\label{tab:arbitrary_length}
\end{table}

\section{Random Seed Robustness Test}
\label{sec:random_seed_test}

Unless otherwise specified, all experiments in this paper use the same fixed random seed. To assess the robustness of \method{} to random seed selection, we additionally evaluate \method{}-IS using five different random seeds on ETTh1, ETTm1, and Weather. Table~\ref{tab:random_seed} reports the standard deviation and average MSE/MAE across seeds, together with the results of SOTA baselines. ETTh1, a relatively small dataset, exhibits larger standard deviations than ETTm1 and Weather. Nevertheless, its average MSE and MAE across the five seeds remain lower than those of the SOTA baselines.

\begin{table*}[ht]
\caption{Forecasting performance across five random seeds ($L=96$, averaged over output lengths $T \in \{96,192,336,720\}$). \textbf{Bold} and \underline{underlined} values indicate the best and second-best results when comparing \method{}-IS AVG with the baselines.}
\centering
\scriptsize
\setlength{\tabcolsep}{1pt}
\renewcommand{\arraystretch}{0.5}
\resizebox{\textwidth}{!}{%
\begin{tabular}{l | S[table-format=1.3] S[table-format=1.3] | >{\columncolor{green!15}}S[table-format=1.3] >{\columncolor{green!15}}S[table-format=1.3] | S[table-format=1.3] S[table-format=1.3] | S[table-format=1.3] S[table-format=1.3] | S[table-format=1.3] S[table-format=1.3] | S[table-format=1.3] S[table-format=1.3]}
\toprule
& \multicolumn{2}{c|}{\shortstack{\method{}-IS\\STD}} & \multicolumn{2}{c|}{\cellcolor{green!15}\shortstack{\method{}-IS\\AVG}} & \multicolumn{2}{c|}{\shortstack{DGraFormer\\(2025)}} & \multicolumn{2}{c|}{\shortstack{CASA\\(2025)}} & \multicolumn{2}{c|}{\shortstack{TQNet\\(2025)}} & \multicolumn{2}{c}{\shortstack{S-Mamba\\(2025)}} \\
\cmidrule(lr){2-3}\cmidrule(lr){4-5}\cmidrule(lr){6-7}\cmidrule(lr){8-9}\cmidrule(lr){10-11}\cmidrule(lr){12-13}
Dataset & {MSE} & {MAE} & {MSE} & {MAE} & {MSE} & {MAE} & {MSE} & {MAE} & {MSE} & {MAE} & {MSE} & {MAE} \\
\midrule
ETTh1 & 0.008 & 0.007 & \textbf{0.432} & \textbf{0.428} & 0.442 & \underline{0.432} & \underline{0.438} & 0.433 & 0.441 & 0.434 & 0.455 & 0.450 \\
\midrule
ETTm1 & 0.001 & 0.001 & \textbf{0.366} & \textbf{0.379} & 0.379 & 0.394 & 0.386 & 0.394 & \underline{0.377} & \underline{0.393} & 0.398 & 0.405 \\
\midrule
Weather & 0.001 & 0.001 & \textbf{0.234} & \textbf{0.262} & 0.249 & 0.271 & 0.243 & \underline{0.267} & \underline{0.242} & 0.269 & 0.251 & 0.276 \\
\bottomrule
\end{tabular}%
}
\label{tab:random_seed}
\end{table*}

\newpage
\clearpage
\section{Full Experimental Results for Fixed-Input Forecasting}
\label{sec:Full_FIFO_FIMO}

Table~\ref{tab:fixdynamic} summarizes the forecasting performance across six benchmark datasets. By comparing the horizon-specific \method{}-IS against the horizon-unified \method{}-HU and current SOTA baselines, we draw two critical conclusions.

\paragraph{Superiority of the Segment Tree Architecture.}
In the standard fixed-horizon setting (\textsf{S1}), \method{}-IS establishes a new SOTA, achieving the lowest average MSE on 5 out of 6 datasets (the second-best in the remaining dataset). Notably, on {ETTh1} and {ETTm1}, \method{}-IS outperforms complex Transformer-based models like DGraFormer and TQNet. 
This validates that our Segment Tree Encoder captures multi-scale temporal dependencies more effectively than fixed-window attention mechanisms, while the Pascal-based Range Weighter acts as a precise filter to suppress noise in volatile series.

\paragraph{Effectiveness of Parameter Decoupling (\textsf{S1} vs. \textsf{S2}).}
A key contribution of this work is the ability to decouple model parameters from output length $T$. The \method{}-HU variant, which optimizes a single set of parameters for all four horizons simultaneously (\textsf{S2}), demonstrates remarkable resilience. As shown in Table~\ref{tab:fixdynamic}, the performance gap between the specialized \method{}-IS and the unified \method{}-HU is negligible --- often less than 0.005 in MSE. 
In specific cases like {ETTm1}, \method{}-HU even marginally outperforms the specialized models. This confirms that the Inverse Tree Decoder successfully generalizes across planning horizons without the ``catastrophic forgetting'' or parameter dilution typically observed when traditional models are forced to handle variable outputs with heuristic adapters.


\begin{table*}[ht]
\caption{Multivariate forecasting performance ($L=96$, $T \in \{96, 192, 336, 720\}$). We compare {\method{}-IS} and all baselines (trained specifically for each horizon, Setting \textsf{S1}) against {\method{}-HU} (a single unified model trained for all horizons, Setting \textsf{S2}). \textbf{Bold} and \underline{underlined} values indicate the best and second-best results, respectively.}
\centering
\scriptsize
\setlength{\tabcolsep}{1pt}
\renewcommand{\arraystretch}{1.3}
\resizebox{\textwidth}{!}{%
\begin{tabular}{l c | >{\columncolor{green!15}}S[table-format=1.3] >{\columncolor{green!15}}S[table-format=1.3] | >{\columncolor{green!15}}S[table-format=1.3] >{\columncolor{green!15}}S[table-format=1.3] | S[table-format=1.3] S[table-format=1.3] | S[table-format=1.3] S[table-format=1.3] | S[table-format=1.3] S[table-format=1.3] | S[table-format=1.3] S[table-format=1.3] | S[table-format=1.3] S[table-format=1.3] | S[table-format=1.3] S[table-format=1.3] | S[table-format=1.3] S[table-format=1.3] | S[table-format=1.3] S[table-format=1.3] | S[table-format=1.3] S[table-format=1.3] | S[table-format=1.3] S[table-format=1.3] | S[table-format=1.3] S[table-format=1.3]}
\toprule
& & \multicolumn{2}{c|}{\cellcolor{green!15}\shortstack{\method{}\\-IS (\textsf{S1})}} & \multicolumn{2}{c|}{\cellcolor{green!15}\shortstack{\method{}\\-HU (\textsf{S2})}} & \multicolumn{2}{c|}{\shortstack{DGraF.\\(2025)}} & \multicolumn{2}{c|}{\shortstack{CASA\\(2025)}} & \multicolumn{2}{c|}{\shortstack{TQNet\\(2025)}} & \multicolumn{2}{c|}{\shortstack{S-Mamba\\(2025)}} & \multicolumn{2}{c|}{\shortstack{SOFTS\\(2024)}} & \multicolumn{2}{c|}{\shortstack{iTrans.\\(2024)}} & \multicolumn{2}{c|}{\shortstack{PatchTST\\(2023)}} & \multicolumn{2}{c|}{\shortstack{TSMixer\\(2023)}} & \multicolumn{2}{c|}{\shortstack{TiDE\\(2023)}} & \multicolumn{2}{c|}{\shortstack{DLinear\\(2023)}} & \multicolumn{2}{c}{\shortstack{LSTM\\(1997)}} \\
\cmidrule(lr){3-4}\cmidrule(lr){5-6}\cmidrule(lr){7-8}\cmidrule(lr){9-10}\cmidrule(lr){11-12}\cmidrule(lr){13-14}\cmidrule(lr){15-16}\cmidrule(lr){17-18}\cmidrule(lr){19-20}\cmidrule(lr){21-22}\cmidrule(lr){23-24}\cmidrule(lr){25-26}\cmidrule(lr){27-28}
Dataset & T & {MSE} & {MAE} & {MSE} & {MAE} & {MSE} & {MAE} & {MSE} & {MAE} & {MSE} & {MAE} & {MSE} & {MAE} & {MSE} & {MAE} & {MSE} & {MAE} & {MSE} & {MAE} & {MSE} & {MAE} & {MSE} & {MAE} & {MSE} & {MAE} & {MSE} & {MAE} \\
\midrule
\multirow{5}{*}{ETTm1}
& 96 & \underline{0.303} & \underline{0.337} & \textbf{0.300} & \textbf{0.336} & 0.326 & 0.366 & 0.313 & 0.358 & 0.311 & 0.353 & 0.333 & 0.368 & 0.325 & 0.361 & 0.334 & 0.368 & 0.329 & 0.365 & 0.323 & 0.363 & 0.364 & 0.387 & 0.345 & 0.372 & 1.010 & 0.764 \\
& 192 & \underline{0.355} & \underline{0.366} & \textbf{0.350} & \textbf{0.365} & 0.358 & 0.376 & 0.375 & 0.382 & 0.356 & 0.378 & 0.376 & 0.390 & 0.377 & 0.391 & 0.380 & 0.394 & 0.376 & 0.392 & 0.450 & 0.451 & 0.374 & 0.387 & 0.380 & 0.389 & 1.221 & 0.851 \\
& 336 & \underline{0.381} & \underline{0.388} & \textbf{0.377} & \textbf{0.387} & 0.387 & 0.397 & 0.397 & 0.398 & 0.390 & 0.401 & 0.408 & 0.413 & 0.426 & 0.420 & 0.400 & 0.410 & 0.407 & 0.413 & 0.532 & 0.515 & 0.410 & 0.411 & 0.413 & 0.413 & 1.182 & 0.849 \\
& 720 & \textbf{0.422} & \underline{0.424} & \underline{0.436} & \textbf{0.421} & 0.446 & 0.436 & 0.459 & 0.437 & 0.452 & 0.440 & 0.475 & 0.448 & 0.466 & 0.447 & 0.491 & 0.459 & 0.475 & 0.453 & 0.485 & 0.459 & 0.487 & 0.461 & 0.474 & 0.453 & 1.312 & 0.835 \\
\cmidrule(lr){2-28}
& Avg & \textbf{0.365} & \underline{0.379} & \underline{0.366} & \textbf{0.377} & 0.379 & 0.394 & 0.386 & 0.394 & 0.377 & 0.393 & 0.398 & 0.405 & 0.399 & 0.405 & 0.401 & 0.408 & 0.397 & 0.406 & 0.448 & 0.447 & 0.409 & 0.411 & 0.403 & 0.407 & 1.181 & 0.825 \\
\midrule
\multirow{5}{*}{ETTm2}
& 96 & \textbf{0.168} & \textbf{0.249} & \underline{0.173} & 0.254 & \underline{0.173} & 0.253 & \underline{0.173} & \underline{0.252} & \underline{0.173} & 0.256 & 0.179 & 0.263 & 0.180 & 0.261 & 0.180 & 0.264 & 0.184 & 0.264 & 0.182 & 0.266 & 0.207 & 0.305 & 0.193 & 0.292 & 1.699 & 0.996 \\
& 192 & \textbf{0.234} & \textbf{0.295} & \underline{0.237} & \underline{0.296} & 0.243 & 0.302 & 0.240 & 0.298 & 0.238 & 0.298 & 0.250 & 0.309 & 0.246 & 0.306 & 0.250 & 0.309 & 0.246 & 0.306 & 0.249 & 0.309 & 0.290 & 0.364 & 0.284 & 0.362 & 2.492 & 1.210 \\
& 336 & \textbf{0.292} & \textbf{0.331} & \underline{0.296} & \underline{0.334} & 0.315 & 0.348 & \underline{0.296} & \underline{0.334} & 0.301 & 0.340 & 0.312 & 0.349 & 0.319 & 0.352 & 0.311 & 0.348 & 0.308 & 0.346 & 0.309 & 0.347 & 0.377 & 0.422 & 0.369 & 0.427 & 2.477 & 1.205 \\
& 720 & \textbf{0.392} & \textbf{0.389} & \underline{0.395} & \underline{0.392} & 0.404 & 0.401 & \underline{0.395} & \underline{0.392} & 0.397 & 0.396 & 0.411 & 0.406 & 0.405 & 0.401 & 0.412 & 0.407 & 0.409 & 0.402 & 0.416 & 0.408 & 0.558 & 0.524 & 0.554 & 0.522 & 3.581 & 1.635 \\
\cmidrule(lr){2-28}
& Avg & \textbf{0.272} & \textbf{0.316} & \underline{0.275} & \underline{0.319} & 0.284 & 0.326 & 0.276 & \underline{0.319} & 0.277 & 0.323 & 0.288 & 0.332 & 0.287 & 0.330 & 0.288 & 0.332 & 0.287 & 0.330 & 0.289 & 0.333 & 0.358 & 0.404 & 0.350 & 0.401 & 2.562 & 1.262 \\
\midrule
\multirow{5}{*}{ETTh1}
& 96 & \underline{0.369} & \textbf{0.385} & 0.372 & \underline{0.391} & \textbf{0.367} & 0.393 & 0.376 & 0.397 & 0.371 & 0.393 & 0.386 & 0.405 & 0.381 & 0.399 & 0.386 & 0.405 & 0.394 & 0.406 & 0.401 & 0.412 & 0.479 & 0.464 & 0.386 & 0.400 & 1.150 & 0.831 \\
& 192 & \textbf{0.419} & \textbf{0.418} & \underline{0.422} & \underline{0.420} & 0.438 & 0.426 & 0.427 & 0.424 & 0.428 & 0.426 & 0.443 & 0.437 & 0.435 & 0.431 & 0.441 & 0.436 & 0.440 & 0.435 & 0.452 & 0.442 & 0.525 & 0.492 & 0.437 & 0.432 & 1.263 & 0.878 \\
& 336 & \underline{0.465} & \textbf{0.440} & \textbf{0.462} & \textbf{0.440} & 0.479 & 0.442 & 0.469 & 0.445 & 0.476 & 0.446 & 0.489 & 0.468 & 0.480 & 0.452 & 0.487 & 0.458 & 0.491 & 0.462 & 0.492 & 0.463 & 0.565 & 0.515 & 0.481 & 0.459 & 1.237 & 0.822 \\
& 720 & \textbf{0.458} & \textbf{0.456} & \textbf{0.458} & \textbf{0.456} & 0.484 & 0.467 & 0.479 & 0.468 & 0.487 & 0.470 & 0.502 & 0.489 & 0.499 & 0.488 & 0.503 & 0.491 & 0.487 & 0.479 & 0.507 & 0.490 & 0.594 & 0.558 & 0.519 & 0.516 & 1.200 & 0.841 \\
\cmidrule(lr){2-28}
& Avg & \textbf{0.428} & \textbf{0.425} & \textbf{0.428} & \underline{0.427} & 0.442 & 0.432 & 0.438 & 0.433 & 0.441 & 0.434 & 0.455 & 0.450 & 0.449 & 0.443 & 0.454 & 0.448 & 0.453 & 0.446 & 0.463 & 0.452 & 0.541 & 0.507 & 0.456 & 0.452 & 1.212 & 0.843 \\
\midrule
\multirow{5}{*}{ETTh2}
& 96 & \textbf{0.283} & \textbf{0.335} & 0.289 & \underline{0.337} & 0.300 & 0.344 & 0.290 & 0.339 & 0.295 & 0.343 & 0.296 & 0.348 & 0.297 & 0.347 & 0.297 & 0.349 & \underline{0.288} & 0.340 & 0.319 & 0.361 & 0.400 & 0.440 & 0.333 & 0.387 & 3.363 & 1.419 \\
& 192 & \textbf{0.366} & \textbf{0.386} & 0.372 & \underline{0.388} & 0.382 & 0.397 & \textbf{0.366} & \underline{0.388} & 0.367 & 0.393 & 0.376 & 0.396 & 0.373 & 0.394 & 0.380 & 0.400 & 0.376 & 0.395 & 0.402 & 0.410 & 0.528 & 0.509 & 0.477 & 0.476 & 3.438 & 1.417 \\
& 336 & \underline{0.407} & \underline{0.418} & 0.409 & 0.422 & \textbf{0.388} & \textbf{0.415} & 0.416 & 0.425 & 0.417 & 0.427 & 0.424 & 0.431 & 0.410 & 0.426 & 0.428 & 0.432 & 0.440 & 0.451 & 0.444 & 0.446 & 0.643 & 0.571 & 0.594 & 0.541 & 2.971 & 1.339 \\
& 720 & \textbf{0.410} & \underline{0.435} & 0.417 & 0.437 & 0.418 & 0.436 & 0.420 & 0.439 & 0.433 & 0.446 & 0.426 & 0.444 & \underline{0.411} & \textbf{0.433} & 0.427 & 0.445 & 0.436 & 0.453 & 0.441 & 0.450 & 0.874 & 0.679 & 0.831 & 0.657 & 3.359 & 1.549 \\
\cmidrule(lr){2-28}
& Avg & \textbf{0.366} & \textbf{0.394} & \underline{0.372} & \underline{0.396} & \underline{0.372} & 0.398 & 0.373 & 0.398 & 0.378 & 0.402 & 0.381 & 0.405 & 0.373 & 0.400 & 0.383 & 0.406 & 0.385 & 0.410 & 0.402 & 0.417 & 0.611 & 0.550 & 0.559 & 0.515 & 3.283 & 1.431 \\
\midrule
\multirow{5}{*}{Weather}
& 96 & \textbf{0.149} & \textbf{0.187} & \underline{0.151} & \underline{0.193} & 0.168 & 0.207 & 0.155 & 0.197 & 0.157 & 0.200 & 0.165 & 0.210 & 0.166 & 0.208 & 0.174 & 0.214 & 0.176 & 0.217 & 0.166 & 0.210 & 0.202 & 0.261 & 0.196 & 0.255 & 0.263 & 0.296 \\
& 192 & \textbf{0.200} & \textbf{0.240} & \underline{0.204} & \underline{0.244} & 0.213 & 0.249 & 0.206 & 0.245 & 0.206 & 0.245 & 0.214 & 0.252 & 0.217 & 0.253 & 0.221 & 0.254 & 0.221 & 0.256 & 0.215 & 0.256 & 0.242 & 0.298 & 0.237 & 0.296 & 0.348 & 0.352 \\
& 336 & \textbf{0.258} & \textbf{0.285} & \underline{0.261} & 0.287 & 0.270 & 0.291 & 0.265 & \underline{0.286} & 0.262 & 0.287 & 0.274 & 0.297 & 0.282 & 0.300 & 0.278 & 0.296 & 0.275 & 0.296 & 0.287 & 0.300 & 0.287 & 0.335 & 0.283 & 0.335 & 0.425 & 0.415 \\
& 720 & \textbf{0.329} & \textbf{0.333} & \underline{0.339} & \underline{0.337} & 0.345 & 0.338 & 0.347 & 0.340 & 0.344 & 0.342 & 0.350 & 0.345 & 0.356 & 0.351 & 0.358 & 0.347 & 0.352 & 0.346 & 0.355 & 0.348 & 0.351 & 0.386 & 0.345 & 0.381 & 0.587 & 0.541 \\
\cmidrule(lr){2-28}
& Avg & \textbf{0.234} & \textbf{0.261} & \underline{0.239} & \underline{0.265} & 0.249 & 0.271 & 0.243 & 0.267 & 0.242 & 0.269 & 0.251 & 0.276 & 0.255 & 0.278 & 0.258 & 0.278 & 0.256 & 0.279 & 0.256 & 0.278 & 0.270 & 0.320 & 0.265 & 0.317 & 0.406 & 0.401 \\
\midrule
\multirow{5}{*}{Solar}
& 96 & \textbf{0.162} & \textbf{0.214} & 0.175 & 0.223 & 0.184 & 0.219 & 0.187 & \underline{0.218} & \underline{0.173} & 0.233 & 0.205 & 0.244 & 0.200 & 0.230 & 0.203 & 0.237 & 0.205 & 0.246 & 0.221 & 0.275 & 0.312 & 0.399 & 0.290 & 0.378 & 0.314 & 0.266 \\
& 192 & \textbf{0.196} & \underline{0.235} & 0.200 & 0.240 & 0.211 & \textbf{0.234} & 0.223 & 0.243 & \underline{0.199} & 0.257 & 0.237 & 0.270 & 0.229 & 0.253 & 0.233 & 0.261 & 0.237 & 0.267 & 0.268 & 0.306 & 0.339 & 0.416 & 0.320 & 0.398 & 0.235 & 0.245 \\
& 336 & 0.231 & 0.256 & \underline{0.218} & \textbf{0.247} & 0.234 & \underline{0.250} & 0.237 & 0.257 & \textbf{0.211} & 0.263 & 0.258 & 0.288 & 0.243 & 0.269 & 0.248 & 0.273 & 0.250 & 0.276 & 0.272 & 0.294 & 0.368 & 0.430 & 0.353 & 0.415 & 0.670 & 0.615 \\
& 720 & \underline{0.227} & \textbf{0.252} & 0.235 & 0.257 & 0.235 & \underline{0.254} & 0.239 & 0.258 & \textbf{0.209} & 0.270 & 0.260 & 0.288 & 0.245 & 0.272 & 0.249 & 0.275 & 0.252 & 0.275 & 0.281 & 0.313 & 0.370 & 0.425 & 0.356 & 0.413 & 0.773 & 0.620 \\
\cmidrule(lr){2-28}
& Avg & \underline{0.204} & \textbf{0.239} & 0.207 & 0.242 & 0.216 & \textbf{0.239} & 0.222 & 0.244 & \textbf{0.198} & 0.256 & 0.240 & 0.272 & 0.229 & 0.256 & 0.233 & 0.262 & 0.236 & 0.266 & 0.261 & 0.297 & 0.347 & 0.417 & 0.330 & 0.401 & 0.498 & 0.436 \\
\bottomrule
\end{tabular}%
}
\label{tab:fixdynamic}
\end{table*}

\newpage
\section{Full MIOL (Multiple Input and Output Lengths) Scenario}
\label{sec:Full_MIOL}

In this section, we validate the model's adaptability in dynamic deployment environments where both lookback windows $L$ and forecast horizons $T$ vary during inference ($L, T \in \{48, 96\}$).

\subsection{Baselines and Heuristic Adapters}
\label{ssec:heuristic_adapters_baselines}

To rigorously evaluate \method{}, we compare it against two baseline categories: the idealized specialist and standard heuristic adaptations.
We use \method{}-MIOL to refer to our native unified model that support multiple input and output lengths.

\paragraph{1. Idealized Specialist (\method{}-IS).}
This represents a theoretical performance upper bound, where a separate, dedicated a model is trained for every distinct $(L, T)$ configuration.

\paragraph{2. Heuristic Adapters.}
For fixed-size models lacking native flexibility, we employ heuristic strategies to force variable-length support (illustrated in Figure~\ref{fig:heuristic}).
\begin{itemize}[leftmargin=*]
\item \textbf{Input Adaptation}: \textit{Padding} aligns all sequences to the maximum length $L_{\max}$ by appending zeros (Figure~\ref{fig:heuristic}(a)), while \textit{Trimming} truncates sequences to the minimum length $L_{\min}$ (Figure~\ref{fig:heuristic}(b)).

\begin{figure}[htbp]
\centering\includegraphics[width=0.9\textwidth]{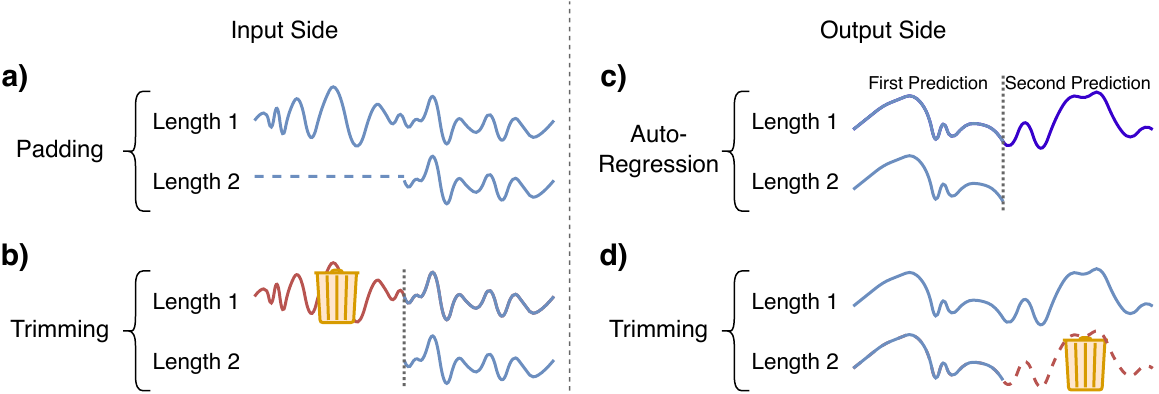}
\caption{Visual illustration of heuristic adaptation strategies. (a) Input Padding; (b) Input Trimming; (c) Output Autoregression; (d) Output Trimming.}
\label{fig:heuristic}
\end{figure}

\item \textbf{Output Adaptation}: \textit{Autoregression} generates the shortest horizon $T_{\min}$ first and iteratively expands to $T_{\max}$ (Figure~\ref{fig:heuristic}(c)). \textit{Trimming} predicts the maximum horizon $T_{\max}$ and discards excess steps to match $T_{\text{target}}$ (Figure~\ref{fig:heuristic}(d)).
\end{itemize}

These strategies are combined to form four baselines: \textsc{Trim}-\textsc{Reg}, \textsc{Trim}-\textsc{Trim}, \textsc{Pad}-\textsc{Reg}, and \textsc{Pad}-\textsc{Trim}.

\subsection{Theoretical Comparison}

As summarized in Table~\ref{tab:miol_comparison}, these heuristics introduce inherent trade-offs. Input padding and output trimming generate computational and memory redundancy ("Computational waste"), while input trimming discards potentially valuable historical context ("Information loss"). Output autoregression, while memory-efficient, suffers from error accumulation and high latency. In contrast, \method{} provides native support without these structural penalties.

\begin{table}[ht]
\centering
\caption{Theoretical comparison of MIOL adaptation strategies versus Native Support.}
\label{tab:miol_comparison}
\begin{small}
\begin{tabular}{lcc}
\toprule
\textbf{Method} & \textbf{Input Side Trade-off} & \textbf{Output Side Trade-off} \\
\midrule
Padding & Computational/Memory Redundancy & -- \\
Trimming & Information Loss & Computational/Memory Redundancy \\
Autoregression & -- & Error Accumulation \& High Latency \\
\midrule
\textbf{\method{} (Ours)} & \textbf{Native Support (Zero Redundancy)} & \textbf{Native Support (Zero Redundancy)} \\
\bottomrule
\end{tabular}
\end{small}
\end{table}

\subsection{Empirical Results and Analysis}
\label{ssec:full_MIOL_res}

Table~\ref{tab:multilen} presents the granular performance results (Accuracy, Latency, Memory) across four input-output length combinations.

\paragraph{Native Elasticity Matches Specialized Training.}
A critical finding is that \method{}-MIOL achieves forecasting accuracy roughly identical to the brute-force \method{}-IS baseline. For example, on {ETTh1}, the average MSE gap between the native MIOL model (0.356) and the specialized IS ensemble (0.353) is negligible. This demonstrates that \method{} successfully learns a length-invariant representation, allowing a single model to replace multiple specialized models without substantial performance degradation.

\paragraph{Superior Efficiency over Padding Heuristics.}
Adapters that rely on padding (e.g., \textsc{Pad}-\textsc{Reg}) incur severe computational penalties. As shown in the {ETTh1} results, the \textsc{Pad}-\textsc{Reg} strategy causes inference latency to spike to 0.633s, nearly \textbf{2.5$\times$ slower} than \method{}-MIOL (0.244s). This confirms that our native variable-length support eliminates the computational redundancy of processing padded zeros, making it significantly more suitable for high-throughput deployment.

\paragraph{Robustness against Trimming Artifacts.}
Adapters that rely on trimming suffer from catastrophic information loss, particularly in volatile datasets. On the {Solar} dataset, the \textsc{Trim}-\textsc{Reg} heuristic results in an MSE of 0.602—a massive degradation compared to \method{}-MIOL's 0.240. This highlights the novelty of our Range Weighter and Feature Fuser: unlike simple trimming, our mechanism intelligently selects the most informative historical ranges, ensuring stability even when input contexts vary dynamically.

\begin{table*}[ht]
\caption{Detailed performance trade-offs in the {Full MIOL Scenario} across four input-output length combinations ($L, T \in \{48, 96\}$). We compare the native {\method{}-MIOL} against the \textcolor{gray}{\method{}-IS} (Idealized Specialist, trained specifically for each $(L,T)$ pair, as the baseline) and four heuristic inference strategies. \emph{Time} and \emph{Mem} denote inference latency (seconds) and memory usage (GB), measured with batch sizes of 2048 (ETT), 1024 (Weather), and 192 (Solar).}
\centering
\scriptsize
\setlength{\tabcolsep}{1pt}
\renewcommand{\arraystretch}{1.1}
\resizebox{\textwidth}{!}{%
\begin{tabular}{l c | >{\color{gray}}S[table-format=1.3] >{\color{gray}}S[table-format=1.3] >{\color{gray}}S[table-format=1.3] >{\color{gray}}S[table-format=2.2] | >{\columncolor{green!15}}S[table-format=1.3] >{\columncolor{green!15}}S[table-format=1.3] >{\columncolor{green!15}}S[table-format=1.3] >{\columncolor{green!15}}S[table-format=2.2] | S[table-format=1.3] S[table-format=1.3] S[table-format=1.3] S[table-format=2.2] | S[table-format=1.3] S[table-format=1.3] S[table-format=1.3] S[table-format=2.2] | S[table-format=1.3] S[table-format=1.3] S[table-format=1.3] S[table-format=2.2] | S[table-format=1.3] S[table-format=1.3] S[table-format=1.3] S[table-format=2.2]}
\toprule
& &
\multicolumn{4}{c|}{\textcolor{gray}{\shortstack{\method{}\\-IS (\textsf{S1})}}} &
\multicolumn{4}{c|}{\cellcolor{green!15}\shortstack{\method{}\\-MIOL (\textsf{S3})}} &
\multicolumn{4}{c|}{\shortstack{\method{}\\+\textsc{Trim}-\textsc{Reg}}} &
\multicolumn{4}{c|}{\shortstack{\method{}\\+\textsc{Trim}-\textsc{Trim}}} &
\multicolumn{4}{c|}{\shortstack{\method{}\\+\textsc{Pad}-\textsc{Reg}}} &
\multicolumn{4}{c}{\shortstack{\method{}\\+\textsc{Pad}-\textsc{Trim}}} \\
\cmidrule(lr){3-6}\cmidrule(lr){7-10}\cmidrule(lr){11-14}\cmidrule(lr){15-18}\cmidrule(lr){19-22}\cmidrule(lr){23-26}
Dataset & $L$/$T$ & {\textcolor{gray}{MSE}} & {\textcolor{gray}{MAE}} & {\textcolor{gray}{Time}} & {\textcolor{gray}{MEM}} & {\cellcolor{green!15}MSE} & {\cellcolor{green!15}MAE} & {\cellcolor{green!15}Time} & {\cellcolor{green!15}MEM} & {MSE} & {MAE} & {Time} & {MEM} & {MSE} & {MAE} & {Time} & {MEM} & {MSE} & {MAE} & {Time} & {MEM} & {MSE} & {MAE} & {Time} & {MEM} \\
\midrule
\multirow{5}{*}{ETTh1}
& 48-48 
& 0.338 & 0.362 & 0.109 & 6.43
& \textbf{0.339} & \textbf{0.363} & 0.106 & 6.47
& 0.347 & \underline{0.367} & 0.110 & 6.48
& \underline{0.341} & 0.370 & 0.124 & 12.74
& 0.367 & 0.394 & 0.380 & 6.68
& 0.351 & 0.386 & 0.475 & 12.90 \\

& 48-96
& 0.376 & 0.389 & 0.111 & 12.63
& \textbf{0.381} & \textbf{0.390} & 0.111 & 12.68
& 0.393 & 0.398 & 0.312 & 6.48
& \underline{0.383} & \underline{0.396} & 0.119 & 12.74
& 0.399 & 0.413 & 0.872 & 6.70
& 0.391 & 0.412 & 0.455 & 12.91 \\

& 96-48
& 0.328 & 0.361 & 0.379 & 6.89
& \underline{0.331} & \underline{0.361} & 0.376 & 6.75
& 0.347 & 0.367 & 0.111 & 6.48
& 0.341 & 0.370 & 0.119 & 12.74
& 0.338 & 0.362 & 0.395 & 6.68
& \textbf{0.326} & \textbf{0.360} & 0.468 & 12.90 \\

& 96-96
& 0.369 & 0.385 & 0.392 & 12.88
& \underline{0.371} & \underline{0.387} & 0.383 & 12.95
& 0.393 & 0.398 & 0.305 & 6.49
& 0.383 & 0.396 & 0.120 & 12.75
& 0.375 & 0.388 & 0.884 & 6.70
& \textbf{0.369} & \textbf{0.386} & 0.460 & 12.91 \\
\cmidrule(lr){2-26}
& AVG
& 0.353 & 0.374 & 0.248 & 9.71
& \textbf{0.356} & \textbf{0.375} & 0.244 & 9.71
& 0.370 & \underline{0.383} & 0.210 & 6.48
& 0.362 & \underline{0.383} & 0.121 & 12.74
& 0.370 & 0.389 & 0.633 & 6.69
& \underline{0.359} & 0.386 & 0.465 & 12.91 \\
\midrule
\multirow{5}{*}{ETTh2}
& 48-48
& 0.230 & 0.294 & 0.114 & 6.47
& \underline{0.230} & \underline{0.296} & 0.110 & 6.44
& 0.234 & 0.300 & 0.108 & 6.44
& 0.231 & \textbf{0.294} & 0.111 & 12.65
& \textbf{0.225} & 0.310 & 0.395 & 6.90
& 0.231 & 0.304 & 0.408 & 12.95 \\

& 48-96
& 0.291 & 0.336 & 0.108 & 12.67
& \underline{0.298} & \underline{0.340} & 0.111 & 12.64
& 0.301 & 0.344 & 0.311 & 6.45
& 0.299 & \textbf{0.338} & 0.113 & 12.65
& \textbf{0.288} & 0.354 & 0.864 & 6.91
& 0.302 & 0.353 & 0.409 & 12.96 \\

& 96-48
& 0.220 & 0.291 & 0.375 & 6.68
& \textbf{0.218} & \underline{0.290} & 0.395 & 6.68
& 0.234 & 0.300 & 0.108 & 6.44
& 0.231 & 0.294 & 0.113 & 12.65
& 0.234 & 0.305 & 0.382 & 6.89
& \underline{0.220} & \textbf{0.288} & 0.420 & 12.95 \\

& 96-96
& 0.283 & 0.335 & 0.447 & 12.91
& \underline{0.285} & \underline{0.334} & 0.411 & 12.88
& 0.301 & 0.344 & 0.329 & 6.45
& 0.299 & 0.338 & 0.113 & 12.65
& 0.304 & 0.354 & 0.870 & 6.91
& \textbf{0.283} & \textbf{0.330} & 0.417 & 12.95 \\
\cmidrule(lr){2-26}
& AVG
& 0.256 & 0.314 & 0.261 & 9.68
& \textbf{0.258} & \textbf{0.315} & 0.257 & 9.66
& 0.268 & 0.322 & 0.214 & 6.45
& 0.265 & \underline{0.316} & 0.113 & 12.65
& 0.263 & 0.331 & 0.628 & 6.90
& \underline{0.259} & 0.319 & 0.414 & 12.95 \\
\midrule
\multirow{5}{*}{ETTm1}
& 48-48
& 0.392 & 0.383 & 0.105 & 6.47
& \textbf{0.412} & \textbf{0.388} & 0.131 & 6.45
& 0.456 & 0.411 & 0.125 & 6.48
& \underline{0.419} & \underline{0.393} & 0.109 & 12.64
& 0.530 & 0.455 & 0.457 & 6.72
& 0.462 & 0.419 & 0.405 & 12.91 \\

& 48-96
& 0.382 & 0.380 & 0.120 & 12.65
& \textbf{0.379} & \textbf{0.380} & 0.131 & 12.65
& 0.436 & 0.413 & 0.338 & 6.49
& \underline{0.387} & \underline{0.385} & 0.109 & 12.64
& 0.459 & 0.423 & 0.999 & 6.73
& 0.428 & 0.411 & 0.399 & 12.92 \\

& 96-48
& 0.269 & 0.314 & 0.409 & 6.68
& \textbf{0.273} & \textbf{0.317} & 0.418 & 6.69
& 0.456 & 0.411 & 0.122 & 6.48
& 0.419 & 0.393 & 0.109 & 12.64
& 0.304 & 0.345 & 0.447 & 6.71
& \underline{0.278} & \underline{0.322} & 0.400 & 12.91 \\

& 96-96
& 0.303 & 0.337 & 0.415 & 12.88
& \textbf{0.306} & \textbf{0.341} & 0.401 & 12.89
& 0.436 & 0.413 & 0.337 & 6.49
& 0.387 & 0.385 & 0.111 & 12.65
& 0.336 & 0.364 & 0.977 & 6.73
& \underline{0.310} & \underline{0.344} & 0.408 & 12.91 \\
\cmidrule(lr){2-26}
& AVG
& 0.337 & 0.354 & 0.262 & 9.67
& \textbf{0.343} & \textbf{0.357} & 0.270 & 9.67
& 0.446 & 0.412 & 0.231 & 6.49
& 0.403 & 0.389 & 0.110 & 12.64
& 0.407 & 0.397 & 0.720 & 6.72
& \underline{0.370} & \underline{0.374} & 0.403 & 12.91 \\
\midrule
\multirow{5}{*}{ETTm2}
& 48-48
& 0.137 & 0.226 & 0.112 & 6.54
& \textbf{0.139} & \textbf{0.228} & 0.116 & 6.48
& \underline{0.148} & 0.247 & 0.128 & 6.46
& 0.149 & \underline{0.243} & 0.120 & 12.67
& 0.169 & 0.277 & 0.428 & 6.76
& 0.158 & 0.258 & 0.380 & 12.91 \\

& 48-96
& 0.179 & 0.255 & 0.115 & 12.67
& \textbf{0.179} & \textbf{0.255} & 0.120 & 12.68
& \underline{0.186} & 0.270 & 0.319 & 6.47
& 0.188 & \underline{0.268} & 0.120 & 12.68
& 0.192 & 0.281 & 0.955 & 6.78
& 0.204 & 0.296 & 0.396 & 12.92 \\

& 96-48
& 0.128 & 0.219 & 0.380 & 6.89
& \textbf{0.129} & \textbf{0.216} & 0.389 & 6.76
& 0.148 & 0.247 & 0.119 & 6.46
& 0.149 & 0.243 & 0.136 & 12.68
& 0.137 & 0.233 & 0.433 & 6.76
& \underline{0.131} & \underline{0.226} & 0.383 & 12.91 \\

& 96-96
& 0.168 & 0.249 & 0.395 & 12.95
& \textbf{0.171} & \textbf{0.247} & 0.420 & 12.96
& 0.186 & 0.270 & 0.319 & 6.47
& 0.188 & 0.268 & 0.124 & 12.68
& 0.177 & 0.258 & 0.964 & 6.77
& \underline{0.174} & \underline{0.255} & 0.398 & 12.91 \\
\cmidrule(lr){2-26}
& AVG
& 0.153 & 0.237 & 0.251 & 9.76
& \textbf{0.155} & \textbf{0.237} & 0.261 & 9.72
& \underline{0.167} & \underline{0.256} & 0.221 & 6.47
& 0.169 & \underline{0.256} & 0.125 & 12.68
& 0.169 & 0.262 & 0.695 & 6.77
& \underline{0.167} & 0.259 & 0.389 & 12.91 \\
\midrule
\multirow{5}{*}{Weather}
& 48-48 
& 0.140 & 0.176 & 0.104 & 9.31
& \underline{0.142} & \textbf{0.176} & 0.113 & 9.32
& 0.171 & 0.204 & 0.119 & 9.38
& 0.148 & \underline{0.182} & 0.125 & 18.47
& \textbf{0.140} & \underline{0.182} & 0.413 & 9.47
& 0.144 & 0.183 & 0.395 & 18.64 \\

& 48-96
& 0.172 & 0.214 & 0.108 & 18.46
& \underline{0.178} & \textbf{0.218} & 0.114 & 18.48
& 0.214 & 0.254 & 0.366 & 9.39
& 0.188 & 0.229 & 0.112 & 18.48
& \textbf{0.175} & \underline{0.222} & 0.933 & 9.49
& 0.180 & 0.223 & 0.391 & 18.65 \\

& 96-48
& 0.119 & 0.151 & 0.424 & 9.44
& \underline{0.119} & \underline{0.153} & 0.398 & 9.45
& 0.171 & 0.204 & 0.117 & 9.38
& 0.148 & 0.182 & 0.113 & 18.48
& 0.120 & 0.156 & 0.413 & 9.47
& \textbf{0.114} & \textbf{0.146} & 0.407 & 18.64 \\

& 96-96
& 0.149 & 0.187 & 0.395 & 18.59
& \underline{0.150} & \underline{0.190} & 0.383 & 18.61
& 0.214 & 0.254 & 0.366 & 9.39
& 0.188 & 0.229 & 0.112 & 18.48
& 0.152 & 0.195 & 0.928 & 9.49
& \textbf{0.145} & \textbf{0.185} & 0.406 & 18.64 \\
\cmidrule(lr){2-26}
& AVG
& 0.145 & 0.182 & 0.258 & 13.95
& \underline{0.147} & \textbf{0.184} & 0.252 & 13.97
& 0.192 & 0.229 & 0.242 & 9.39
& 0.168 & 0.205 & 0.116 & 18.48
& \underline{0.147} & 0.189 & 0.672 & 9.48
& \textbf{0.146} & \textbf{0.184} & 0.400 & 18.64 \\
\midrule
\multirow{5}{*}{Solar}
& 48-48
& 0.325 & 0.361 & 0.127 & 11.24
& 0.297 & 0.330 & 0.117 & 11.33
& 0.546 & 0.565 & 0.117 & 11.33
& \textbf{0.283} & 0.318 & 0.122 & 22.45
& \underline{0.285} & \underline{0.304} & 0.416 & 11.37
& 0.295 & \textbf{0.303} & 0.415 & 22.49 \\

& 48-96
& 0.344 & 0.372 & 0.112 & 22.36
& 0.350 & 0.373 & 0.123 & 22.45
& 0.658 & 0.607 & 0.395 & 11.35
& \textbf{0.321} & \underline{0.355} & 0.122 & 22.45
& \underline{0.342} & \textbf{0.344} & 0.954 & 11.40
& 0.361 & 0.358 & 0.401 & 22.50 \\

& 96-48
& 0.139 & 0.198 & 0.401 & 11.28
& \textbf{0.144} & \textbf{0.210} & 0.417 & 11.37
& 0.546 & 0.565 & 0.117 & 11.34
& 0.283 & 0.318 & 0.122 & 22.45
& 0.160 & 0.242 & 0.415 & 11.37
& \underline{0.148} & \underline{0.222} & 0.413 & 22.48 \\

& 96-96
& 0.162 & 0.214 & 0.419 & 22.42
& \textbf{0.168} & \textbf{0.225} & 0.414 & 22.49
& 0.658 & 0.607 & 0.398 & 11.36
& 0.321 & 0.355 & 0.122 & 22.46
& 0.188 & 0.274 & 0.954 & 11.40
& \underline{0.172} & \underline{0.239} & 0.417 & 22.49 \\
\cmidrule(lr){2-26}
& AVG
& 0.243 & 0.286 & 0.265 & 16.83
& \textbf{0.240} & \underline{0.285} & 0.268 & 16.91
& 0.602 & 0.586 & 0.257 & 11.34
& 0.302 & 0.336 & 0.122 & 22.45
& \underline{0.244} & 0.291 & 0.685 & 11.39
& \underline{0.244} & \textbf{0.281} & 0.412 & 22.49 \\
\bottomrule
\end{tabular}%
}
\label{tab:multilen}
\end{table*}

To fully demonstrate \method's robust MIOL capability under challenging length ranges, we expanded the combinations from 4 to 8 pairs including standard long-term horizons:
\begin{equation}
L \in \{48, 96, 144\}, \quad T \in 
\begin{cases}
\{96, 192\}, & L = 48 \\
\{96, 192, 336, 720\}, & L = 96 \\
\{336, 720\}, & L = 144
\end{cases}.
\end{equation}
As shown in Table~\ref{tab:new_multilen}, despite the immensely increased task difficulty, \method consistently preserves peak performance and occasionally outperforms the \method-ISs which are independently trained models.

\begin{table*}[ht]
\caption{Multivariate forecasting performance comparing \method-IS (\textsf{S1}) and \method-MIOL (\textsf{S3}) across 8 different $(L, T)$ pairs. \textbf{Bold} indicates the better result.}
\centering
\scriptsize
\setlength{\tabcolsep}{3pt}
\renewcommand{\arraystretch}{1.3}
\begin{tabular}{l c c | S[table-format=1.3] S[table-format=1.3] | S[table-format=1.3] S[table-format=1.3] || l c c | S[table-format=1.3] S[table-format=1.3] | S[table-format=1.3] S[table-format=1.3]}
\toprule
& & & \multicolumn{2}{c|}{\shortstack{\method{}\\-IS (\textsf{S1})}} & \multicolumn{2}{c||}{\shortstack{\method{}\\-MIOL (\textsf{S3})}} & & & & \multicolumn{2}{c|}{\shortstack{\method{}\\-IS (\textsf{S1})}} & \multicolumn{2}{c}{\shortstack{\method{}\\-MIOL (\textsf{S3})}} \\
\cmidrule(lr){4-5}\cmidrule(lr){6-7}\cmidrule(lr){11-12}\cmidrule(lr){13-14}
Dataset & $L$ & $T$ & {MSE} & {MAE} & {MSE} & {MAE} & Dataset & $L$ & $T$ & {MSE} & {MAE} & {MSE} & {MAE} \\
\midrule
\multirow{9}{*}{ETTh1} & 48 & 96 & 0.376 & \textbf{0.389} & \textbf{0.375} & 0.392 & \multirow{9}{*}{ETTm2} & 48 & 96 & \textbf{0.179} & \textbf{0.255} & 0.180 & 0.262 \\
 & 48 & 192 & 0.437 & 0.425 & \textbf{0.425} & \textbf{0.422} &  & 48 & 192 & 0.247 & \textbf{0.304} & \textbf{0.246} & 0.305 \\
 & 96 & 96 & 0.369 & \textbf{0.385} & \textbf{0.364} & 0.389 &  & 96 & 96 & \textbf{0.168} & \textbf{0.249} & 0.171 & 0.255 \\
 & 96 & 192 & 0.419 & 0.418 & \textbf{0.409} & \textbf{0.417} &  & 96 & 192 & \textbf{0.234} & \textbf{0.295} & \textbf{0.234} & 0.296 \\
 & 96 & 336 & 0.465 & 0.440 & \textbf{0.438} & \textbf{0.436} &  & 96 & 336 & \textbf{0.292} & \textbf{0.331} & 0.295 & 0.335 \\
 & 96 & 720 & 0.458 & 0.456 & \textbf{0.440} & \textbf{0.454} &  & 96 & 720 & \textbf{0.392} & \textbf{0.389} & 0.398 & 0.395 \\
 & 144 & 336 & 0.458 & 0.435 & \textbf{0.432} & \textbf{0.433} &  & 144 & 336 & \textbf{0.283} & \textbf{0.327} & 0.287 & 0.331 \\
 & 144 & 720 & 0.456 & 0.456 & \textbf{0.439} & \textbf{0.453} &  & 144 & 720 & \textbf{0.376} & \textbf{0.383} & 0.388 & 0.390 \\
\cmidrule(lr){2-7}\cmidrule(lr){9-14}
& \multicolumn{2}{c|}{Avg} & 0.430 & 0.426 & \textbf{0.415} & \textbf{0.425} & & \multicolumn{2}{c|}{Avg} & \textbf{0.271} & \textbf{0.317} & 0.275 & 0.321 \\
\midrule
\multirow{9}{*}{ETTh2} & 48 & 96 & \textbf{0.291} & \textbf{0.336} & 0.303 & 0.346 & \multirow{9}{*}{Weather} & 48 & 96 & \textbf{0.172} & \textbf{0.214} & 0.180 & 0.222 \\
 & 48 & 192 & \textbf{0.381} & \textbf{0.389} & 0.393 & 0.398 &  & 48 & 192 & \textbf{0.220} & \textbf{0.256} & 0.225 & 0.261 \\
 & 96 & 96 & \textbf{0.283} & \textbf{0.335} & 0.290 & 0.339 &  & 96 & 96 & \textbf{0.149} & \textbf{0.187} & 0.152 & 0.194 \\
 & 96 & 192 & \textbf{0.366} & \textbf{0.386} & 0.375 & 0.390 &  & 96 & 192 & \textbf{0.200} & \textbf{0.240} & 0.201 & 0.241 \\
 & 96 & 336 & \textbf{0.407} & \textbf{0.418} & 0.419 & 0.425 &  & 96 & 336 & \textbf{0.258} & 0.285 & \textbf{0.258} & \textbf{0.284} \\
 & 96 & 720 & \textbf{0.410} & \textbf{0.435} & 0.424 & 0.440 &  & 96 & 720 & \textbf{0.329} & \textbf{0.333} & 0.335 & 0.335 \\
 & 144 & 336 & \textbf{0.400} & \textbf{0.417} & 0.405 & 0.418 &  & 144 & 336 & 0.257 & 0.285 & \textbf{0.251} & \textbf{0.279} \\
 & 144 & 720 & \textbf{0.409} & \textbf{0.434} & 0.420 & 0.437 &  & 144 & 720 & 0.328 & 0.333 & \textbf{0.326} & \textbf{0.330} \\
\cmidrule(lr){2-7}\cmidrule(lr){9-14}
& \multicolumn{2}{c|}{Avg} & \textbf{0.368} & \textbf{0.394} & 0.379 & 0.399 & & \multicolumn{2}{c|}{Avg} & \textbf{0.239} & \textbf{0.267} & 0.241 & 0.268 \\
\midrule
\multirow{9}{*}{ETTm1} & 48 & 96 & 0.382 & 0.380 & \textbf{0.380} & \textbf{0.376} & \multirow{9}{*}{Solar} & 48 & 96 & 0.344 & 0.372 & \textbf{0.318} & \textbf{0.352} \\
 & 48 & 192 & \textbf{0.422} & \textbf{0.404} & 0.431 & \textbf{0.404} &  & 48 & 192 & 0.314 & 0.337 & \textbf{0.303} & \textbf{0.329} \\
 & 96 & 96 & 0.303 & \textbf{0.337} & \textbf{0.299} & 0.338 &  & 96 & 96 & \textbf{0.162} & \textbf{0.214} & 0.177 & 0.221 \\
 & 96 & 192 & 0.355 & 0.366 & \textbf{0.343} & \textbf{0.364} &  & 96 & 192 & \textbf{0.196} & \textbf{0.235} & 0.201 & 0.238 \\
 & 96 & 336 & 0.381 & \textbf{0.388} & \textbf{0.376} & \textbf{0.388} &  & 96 & 336 & 0.231 & 0.256 & \textbf{0.223} & \textbf{0.253} \\
 & 96 & 720 & \textbf{0.422} & 0.424 & 0.433 & \textbf{0.421} &  & 96 & 720 & \textbf{0.227} & \textbf{0.252} & 0.241 & 0.262 \\
 & 144 & 336 & 0.380 & 0.387 & \textbf{0.369} & \textbf{0.384} &  & 144 & 336 & 0.225 & \textbf{0.241} & \textbf{0.220} & 0.249 \\
 & 144 & 720 & \textbf{0.421} & 0.422 & 0.422 & \textbf{0.416} &  & 144 & 720 & \textbf{0.224} & \textbf{0.250} & 0.236 & 0.260 \\
\cmidrule(lr){2-7}\cmidrule(lr){9-14}
& \multicolumn{2}{c|}{Avg} & 0.383 & 0.389 & \textbf{0.382} & \textbf{0.386} & & \multicolumn{2}{c|}{Avg} & \textbf{0.240} & \textbf{0.270} & \textbf{0.240} & 0.271 \\
\bottomrule
\end{tabular}
\label{tab:new_multilen}
\end{table*}

\newpage
\section{Full MIOL Scenario for Other TSF Models}
\label{sec:MIOL_Other}
\subsection{Benchmarking Against Adapted SOTA Models}
\label{ssec:sota_miol_comparison}

To verify that the limitations of heuristic adapters are not specific to \method{}, we extend our evaluation to recent state-of-the-art architectures, specifically {CASA}~\cite{lee2025casa} and {DGraFormer}~\cite{yan2025dgraformer}. 
Since these models lack native support for variable input/output lengths, we apply the four heuristic strategies defined in Appendix~\ref{sec:Full_MIOL} (e.g., \textsc{Trim}-\textsc{Reg}, \textsc{Pad}-\textsc{Trim}) and quantify their performance stability.

For each TSF model, we first establish an ``IS'' (idealized specialist) baseline by training specialized models for every $(L, T)$ pair to determine optimal theoretical performance. 
We then evaluate the performance using either native support (\method{}-MIOL for \method{}) or heuristic adapters (for SOTA). 
The key metric is the \textbf{relative degradation rate}: the percentage increase in MSE/MAE compared to the specialized -IS baseline, indicating the ``flexibility tax'' incurred by the model.

Table~\ref{tab:multilen_other} reveals a sharp contrast in architectural resilience:
\begin{itemize}[leftmargin=*]
    \item \textbf{Native Stability vs. Heuristic Fragility}:
    \method{}-MIOL demonstrates exceptional stability, typically exhibiting only a minor performance drop compared to its fixed baseline (with an average MSE increase of merely ~0.75\%). In contrast, SOTA models suffer severe degradation under heuristic adaptation. Notably, in some cases, heuristic methods even triggered an MSE surge exceeding +200\%, rendering the models practically unusable for MIOL forecasting.

    \item \textbf{Incompatibility of Rigid Architectures}:
    While models like DGraFormer and CASA perform well in fixed settings, they lack the structural elasticity to handle multiple lengths. Imposing variable lengths on rigid attention mechanisms that rely on precise positional encoding cripples their temporal modeling capability.

    \item \textbf{Architectural Advantage}:
    The results conclusively show that \method{}'s performance stems from its design. By decoupling model parameters from sequence length, \method{} avoids the ``structural mismatch'' that plagues other models when processed via padding or autoregression. This positions \method{} as a robust solution for applications requiring flexible support for multiple input-output lengths (MIOL) configurations.
\end{itemize}

\begin{table*}[ht]
\caption{Comparative analysis of {\method{}-MIOL} (native support) versus SOTA models (CASA, DGraFormer) adapted via heuristics. We evaluate performance across randomized input-output configurations ($L, T \in \{48, 96\}$). The {IS} column represents the ``idealized specialist'' baseline (one model trained per $(L,T)$ pair). For other columns, values in \textcolor{red}{red} indicate performance degradation (MSE/MAE increase) relative to the IS baseline, while \textcolor{blue}{blue} indicates improvement. 
}
\centering
\tiny
\setlength{\tabcolsep}{0.8pt}
\renewcommand{\arraystretch}{1.5}
\resizebox{\textwidth}{!}{%
\begin{tabular}{l c | >{\columncolor{green!15}}S[table-format=1.3] >{\columncolor{green!15}}S[table-format=1.3] | >{\columncolor{green!15}}S[table-format=1.3] >{\columncolor{green!15}}S[table-format=1.3] | S[table-format=1.3] S[table-format=1.3] | S[table-format=1.3] S[table-format=1.3] | S[table-format=1.3] S[table-format=1.3] | S[table-format=1.3] S[table-format=1.3] | S[table-format=1.3] S[table-format=1.3] | S[table-format=1.3] S[table-format=1.3] | S[table-format=1.3] S[table-format=1.3] | S[table-format=1.3] S[table-format=1.3] | S[table-format=1.3] S[table-format=1.3] | S[table-format=1.3] S[table-format=1.3]}
\toprule
& & \multicolumn{4}{c|}{\cellcolor{green!15} \method{}} & \multicolumn{10}{c|}{DGraFormer} & \multicolumn{10}{c}{CASA} \\
\cmidrule(lr){3-6}\cmidrule(lr){7-16}\cmidrule(lr){17-26}
& & \multicolumn{2}{c|}{\cellcolor{green!15} IS (\textsf{S1})} & \multicolumn{2}{c|}{\cellcolor{green!15} MIOL (\textsf{S3})} & \multicolumn{2}{c|}{IS} & \multicolumn{2}{c|}{\textsc{Trim}-\textsc{Reg}} & \multicolumn{2}{c|}{\textsc{Trim}-\textsc{Trim}} & \multicolumn{2}{c|}{\textsc{Pad}-\textsc{Reg}} & \multicolumn{2}{c|}{\textsc{Pad}-\textsc{Trim}} & \multicolumn{2}{c|}{IS} & \multicolumn{2}{c|}{\textsc{Trim}-\textsc{Reg}} & \multicolumn{2}{c|}{\textsc{Trim}-\textsc{Trim}} & \multicolumn{2}{c|}{\textsc{Pad}-\textsc{Reg}} & \multicolumn{2}{c}{\textsc{Pad}-\textsc{Trim}} \\
\cmidrule(lr){3-4}\cmidrule(lr){5-6}\cmidrule(lr){7-8}\cmidrule(lr){9-10}\cmidrule(lr){11-12}\cmidrule(lr){13-14}\cmidrule(lr){15-16}\cmidrule(lr){17-18}\cmidrule(lr){19-20}\cmidrule(lr){21-22}\cmidrule(lr){23-24}\cmidrule(lr){25-26}
Dataset & L/T & {MSE} & {MAE} & {MSE} & {MAE} & {MSE} & {MAE} & {MSE} & {MAE} & {MSE} & {MAE} & {MSE} & {MAE} & {MSE} & {MAE} & {MSE} & {MAE} & {MSE} & {MAE} & {MSE} & {MAE} & {MSE} & {MAE} & {MSE} & {MAE} \\
\midrule
\multirow{10}{*}{ETTh1} & 48-48 & 0.338 & 0.362 & 0.339 & 0.363 & 0.340 & 0.370 & 0.339 & 0.368 & 0.334 & 0.368 & 0.362 & 0.394 & 0.357 & 0.392 & 0.342 & 0.377 & 0.344 & 0.376 & 0.340 & 0.377 & 0.360 & 0.397 & 0.354 & 0.387 \\
 & & {\textcolor{blue}{-}} & {\textcolor{blue}{-}} & {\textcolor{red}{+0.30\%}} & {\textcolor{red}{+0.28\%}} & {\textcolor{blue}{-}} & {\textcolor{blue}{-}} & {\textcolor{blue}{-0.32\%}} & {\textcolor{blue}{-0.44\%}} & {\textcolor{blue}{-1.83\%}} & {\textcolor{blue}{-0.45\%}} & {\textcolor{red}{+6.42\%}} & {\textcolor{red}{+6.60\%}} & {\textcolor{red}{+5.04\%}} & {\textcolor{red}{+6.07\%}} & {\textcolor{blue}{-}} & {\textcolor{blue}{-}} & {\textcolor{red}{+0.66\%}} & {\textcolor{blue}{-0.02\%}} & {\textcolor{blue}{-0.41\%}} & {\textcolor{red}{+0.04\%}} & {\textcolor{red}{+5.48\%}} & {\textcolor{red}{+5.51\%}} & {\textcolor{red}{+3.57\%}} & {\textcolor{red}{+2.80\%}} \\

 & 48-96 & 0.376 & 0.389 & 0.381 & 0.390 & 0.371 & 0.395 & 0.376 & 0.396 & 0.372 & 0.393 & 0.392 & 0.415 & 0.399 & 0.423 & 0.383 & 0.404 & 0.385 & 0.406 & 0.380 & 0.401 & 0.396 & 0.420 & 0.393 & 0.413 \\
 & & {\textcolor{blue}{-}} & {\textcolor{blue}{-}} & {\textcolor{red}{+1.33\%}} & {\textcolor{red}{+0.26\%}} & {\textcolor{blue}{-}} & {\textcolor{blue}{-}} & {\textcolor{red}{+1.18\%}} & {\textcolor{red}{+0.43\%}} & {\textcolor{red}{+0.08\%}} & {\textcolor{blue}{-0.36\%}} & {\textcolor{red}{+5.72\%}} & {\textcolor{red}{+5.09\%}} & {\textcolor{red}{+7.43\%}} & {\textcolor{red}{+7.11\%}} & {\textcolor{blue}{-}} & {\textcolor{blue}{-}} & {\textcolor{red}{+0.39\%}} & {\textcolor{red}{+0.54\%}} & {\textcolor{blue}{-0.90\%}} & {\textcolor{blue}{-0.68\%}} & {\textcolor{red}{+3.34\%}} & {\textcolor{red}{+3.91\%}} & {\textcolor{red}{+2.68\%}} & {\textcolor{red}{+2.20\%}} \\

 & 96-48 & 0.328 & 0.361 & 0.331 & 0.361 & 0.334 & 0.371 & 0.339 & 0.368 & 0.334 & 0.368 & 0.332 & 0.368 & 0.329 & 0.368 & 0.340 & 0.381 & 0.344 & 0.376 & 0.340 & 0.377 & 0.338 & 0.380 & 0.334 & 0.377 \\
 & & {\textcolor{blue}{-}} & {\textcolor{blue}{-}} & {\textcolor{red}{+0.91\%}} & {\textcolor{blue}{0.00\%}} & {\textcolor{blue}{-}} & {\textcolor{blue}{-}} & {\textcolor{red}{+1.44\%}} & {\textcolor{blue}{-0.80\%}} & {\textcolor{blue}{-0.10\%}} & {\textcolor{blue}{-0.81\%}} & {\textcolor{blue}{-0.62\%}} & {\textcolor{blue}{-0.69\%}} & {\textcolor{blue}{-1.50\%}} & {\textcolor{blue}{-0.96\%}} & {\textcolor{blue}{-}} & {\textcolor{blue}{-}} & {\textcolor{red}{+1.08\%}} & {\textcolor{blue}{-1.11\%}} & {\textcolor{red}{+0.01\%}} & {\textcolor{blue}{-1.05\%}} & {\textcolor{blue}{-0.57\%}} & {\textcolor{blue}{-0.24\%}} & {\textcolor{blue}{-1.92\%}} & {\textcolor{blue}{-0.94\%}} \\

 & 96-96 & 0.369 & 0.385 & 0.371 & 0.387 & 0.367 & 0.393 & 0.376 & 0.396 & 0.372 & 0.393 & 0.366 & 0.392 & 0.367 & 0.391 & 0.376 & 0.397 & 0.385 & 0.406 & 0.380 & 0.401 & 0.374 & 0.403 & 0.370 & 0.398 \\
 & & {\textcolor{blue}{-}} & {\textcolor{blue}{-}} & {\textcolor{red}{+0.54\%}} & {\textcolor{red}{+0.52\%}} & {\textcolor{blue}{-}} & {\textcolor{blue}{-}} & {\textcolor{red}{+2.35\%}} & {\textcolor{red}{+0.83\%}} & {\textcolor{red}{+1.23\%}} & {\textcolor{red}{+0.04\%}} & {\textcolor{blue}{-0.30\%}} & {\textcolor{blue}{-0.17\%}} & {\textcolor{red}{+0.01\%}} & {\textcolor{blue}{-0.43\%}} & {\textcolor{blue}{-}} & {\textcolor{blue}{-}} & {\textcolor{red}{+2.31\%}} & {\textcolor{red}{+2.25\%}} & {\textcolor{red}{+1.00\%}} & {\textcolor{red}{+1.00\%}} & {\textcolor{blue}{-0.42\%}} & {\textcolor{red}{+1.47\%}} & {\textcolor{blue}{-1.69\%}} & {\textcolor{red}{+0.34\%}} \\

\cmidrule(lr){2-26}
 & AVG & 0.353 & 0.374 & 0.356 & 0.375 & 0.353 & 0.382 & 0.357 & 0.382 & 0.353 & 0.381 & 0.363 & 0.392 & 0.363 & 0.393 & 0.360 & 0.389 & 0.364 & 0.391 & 0.360 & 0.389 & 0.367 & 0.400 & 0.363 & 0.394 \\
 & & {\textcolor{blue}{-}} & {\textcolor{blue}{-}} & {\textcolor{red}{+0.78\%}} & {\textcolor{red}{+0.27\%}} & {\textcolor{blue}{-}} & {\textcolor{blue}{-}} & {\textcolor{red}{+1.18\%}} & {\textcolor{red}{+0.03\%}} & {\textcolor{blue}{-0.12\%}} & {\textcolor{blue}{-0.39\%}} & {\textcolor{red}{+2.82\%}} & {\textcolor{red}{+2.70\%}} & {\textcolor{red}{+2.81\%}} & {\textcolor{red}{+2.96\%}} & {\textcolor{blue}{-}} & {\textcolor{blue}{-}} & {\textcolor{red}{+1.12\%}} & {\textcolor{red}{+0.44\%}} & {\textcolor{blue}{-0.07\%}} & {\textcolor{blue}{-0.17\%}} & {\textcolor{red}{+1.94\%}} & {\textcolor{red}{+2.66\%}} & {\textcolor{red}{+0.66\%}} & {\textcolor{red}{+1.10\%}} \\
\midrule
\multirow{10}{*}{ETTh2} & 48-48 & 0.230 & 0.294 & 0.230 & 0.296 & 0.234 & 0.297 & 0.234 & 0.298 & 0.236 & 0.298 & 0.248 & 0.318 & 0.246 & 0.316 & 0.243 & 0.306 & 0.244 & 0.307 & 0.245 & 0.307 & 0.324 & 0.378 & 0.291 & 0.360 \\
 & & {\textcolor{blue}{-}} & {\textcolor{blue}{-}} & {\textcolor{blue}{0.00\%}} & {\textcolor{red}{+0.68\%}} & {\textcolor{blue}{-}} & {\textcolor{blue}{-}} & {\textcolor{blue}{-0.25\%}} & {\textcolor{red}{+0.37\%}} & {\textcolor{red}{+0.65\%}} & {\textcolor{red}{+0.39\%}} & {\textcolor{red}{+6.07\%}} & {\textcolor{red}{+7.11\%}} & {\textcolor{red}{+5.25\%}} & {\textcolor{red}{+6.43\%}} & {\textcolor{blue}{-}} & {\textcolor{blue}{-}} & {\textcolor{red}{+0.32\%}} & {\textcolor{red}{+0.63\%}} & {\textcolor{red}{+0.79\%}} & {\textcolor{red}{+0.61\%}} & {\textcolor{red}{+33.27\%}} & {\textcolor{red}{+23.76\%}} & {\textcolor{red}{+19.91\%}} & {\textcolor{red}{+17.69\%}} \\

 & 48-96 & 0.291 & 0.336 & 0.298 & 0.340 & 0.298 & 0.340 & 0.301 & 0.342 & 0.298 & 0.340 & 0.310 & 0.356 & 0.334 & 0.378 & 0.309 & 0.348 & 0.307 & 0.348 & 0.307 & 0.349 & 0.399 & 0.423 & 0.393 & 0.426 \\
 & & {\textcolor{blue}{-}} & {\textcolor{blue}{-}} & {\textcolor{red}{+2.41\%}} & {\textcolor{red}{+1.19\%}} & {\textcolor{blue}{-}} & {\textcolor{blue}{-}} & {\textcolor{red}{+1.10\%}} & {\textcolor{red}{+0.53\%}} & {\textcolor{red}{+0.32\%}} & {\textcolor{red}{+0.03\%}} & {\textcolor{red}{+4.19\%}} & {\textcolor{red}{+4.73\%}} & {\textcolor{red}{+12.43\%}} & {\textcolor{red}{+11.28\%}} & {\textcolor{blue}{-}} & {\textcolor{blue}{-}} & {\textcolor{blue}{-0.60\%}} & {\textcolor{blue}{0.00\%}} & {\textcolor{blue}{-0.40\%}} & {\textcolor{red}{+0.03\%}} & {\textcolor{red}{+29.19\%}} & {\textcolor{red}{+21.28\%}} & {\textcolor{red}{+27.26\%}} & {\textcolor{red}{+22.32\%}} \\

 & 96-48 & 0.220 & 0.291 & 0.218 & 0.290 & 0.223 & 0.294 & 0.234 & 0.298 & 0.236 & 0.298 & 0.225 & 0.295 & 0.224 & 0.296 & 0.238 & 0.309 & 0.244 & 0.307 & 0.245 & 0.307 & 0.237 & 0.308 & 0.238 & 0.309 \\
 & & {\textcolor{blue}{-}} & {\textcolor{blue}{-}} & {\textcolor{blue}{-0.91\%}} & {\textcolor{blue}{-0.34\%}} & {\textcolor{blue}{-}} & {\textcolor{blue}{-}} & {\textcolor{red}{+4.85\%}} & {\textcolor{red}{+1.17\%}} & {\textcolor{red}{+5.79\%}} & {\textcolor{red}{+1.18\%}} & {\textcolor{red}{+1.10\%}} & {\textcolor{red}{+0.06\%}} & {\textcolor{red}{+0.67\%}} & {\textcolor{red}{+0.46\%}} & {\textcolor{blue}{-}} & {\textcolor{blue}{-}} & {\textcolor{red}{+2.48\%}} & {\textcolor{blue}{-0.50\%}} & {\textcolor{red}{+2.96\%}} & {\textcolor{blue}{-0.53\%}} & {\textcolor{blue}{-0.25\%}} & {\textcolor{blue}{-0.41\%}} & {\textcolor{blue}{-0.10\%}} & {\textcolor{red}{+0.11\%}} \\

 & 96-96 & 0.283 & 0.335 & 0.285 & 0.334 & 0.300 & 0.344 & 0.301 & 0.342 & 0.298 & 0.340 & 0.288 & 0.337 & 0.285 & 0.337 & 0.290 & 0.339 & 0.307 & 0.348 & 0.307 & 0.349 & 0.299 & 0.348 & 0.300 & 0.349 \\
 & & {\textcolor{blue}{-}} & {\textcolor{blue}{-}} & {\textcolor{red}{+0.71\%}} & {\textcolor{blue}{-0.30\%}} & {\textcolor{blue}{-}} & {\textcolor{blue}{-}} & {\textcolor{red}{+0.26\%}} & {\textcolor{blue}{-0.62\%}} & {\textcolor{blue}{-0.51\%}} & {\textcolor{blue}{-1.12\%}} & {\textcolor{blue}{-4.04\%}} & {\textcolor{blue}{-2.11\%}} & {\textcolor{blue}{-5.00\%}} & {\textcolor{blue}{-2.16\%}} & {\textcolor{blue}{-}} & {\textcolor{blue}{-}} & {\textcolor{red}{+5.77\%}} & {\textcolor{red}{+2.79\%}} & {\textcolor{red}{+5.98\%}} & {\textcolor{red}{+2.83\%}} & {\textcolor{red}{+3.09\%}} & {\textcolor{red}{+2.73\%}} & {\textcolor{red}{+3.30\%}} & {\textcolor{red}{+2.82\%}} \\

\cmidrule(lr){2-26}
 & AVG & 0.256 & 0.314 & 0.258 & 0.315 & 0.264 & 0.319 & 0.267 & 0.320 & 0.267 & 0.319 & 0.268 & 0.326 & 0.273 & 0.332 & 0.270 & 0.326 & 0.275 & 0.328 & 0.276 & 0.328 & 0.315 & 0.364 & 0.305 & 0.361 \\
 & & {\textcolor{blue}{-}} & {\textcolor{blue}{-}} & {\textcolor{red}{+0.68\%}} & {\textcolor{red}{+0.32\%}} & {\textcolor{blue}{-}} & {\textcolor{blue}{-}} & {\textcolor{red}{+1.35\%}} & {\textcolor{red}{+0.33\%}} & {\textcolor{red}{+1.31\%}} & {\textcolor{red}{+0.07\%}} & {\textcolor{red}{+1.61\%}} & {\textcolor{red}{+2.36\%}} & {\textcolor{red}{+3.39\%}} & {\textcolor{red}{+4.03\%}} & {\textcolor{blue}{-}} & {\textcolor{blue}{-}} & {\textcolor{red}{+2.00\%}} & {\textcolor{red}{+0.76\%}} & {\textcolor{red}{+2.32\%}} & {\textcolor{red}{+0.76\%}} & {\textcolor{red}{+16.61\%}} & {\textcolor{red}{+11.88\%}} & {\textcolor{red}{+13.14\%}} & {\textcolor{red}{+10.88\%}} \\
\midrule
\multirow{10}{*}{ETTm1} & 48-48 & 0.392 & 0.383 & 0.412 & 0.388 & 0.463 & 0.419 & 0.450 & 0.412 & 0.441 & 0.405 & 0.557 & 0.462 & 0.545 & 0.456 & 0.429 & 0.408 & 0.421 & 0.402 & 0.425 & 0.407 & 0.461 & 0.418 & 0.465 & 0.425 \\
 & & {\textcolor{blue}{-}} & {\textcolor{blue}{-}} & {\textcolor{red}{+5.10\%}} & {\textcolor{red}{+1.31\%}} & {\textcolor{blue}{-}} & {\textcolor{blue}{-}} & {\textcolor{blue}{-2.78\%}} & {\textcolor{blue}{-1.68\%}} & {\textcolor{blue}{-4.71\%}} & {\textcolor{blue}{-3.28\%}} & {\textcolor{red}{+20.34\%}} & {\textcolor{red}{+10.16\%}} & {\textcolor{red}{+17.68\%}} & {\textcolor{red}{+8.81\%}} & {\textcolor{blue}{-}} & {\textcolor{blue}{-}} & {\textcolor{blue}{-1.88\%}} & {\textcolor{blue}{-1.64\%}} & {\textcolor{blue}{-0.98\%}} & {\textcolor{blue}{-0.48\%}} & {\textcolor{red}{+7.46\%}} & {\textcolor{red}{+2.39\%}} & {\textcolor{red}{+8.18\%}} & {\textcolor{red}{+3.94\%}} \\

 & 48-96 & 0.382 & 0.380 & 0.379 & 0.380 & 0.403 & 0.393 & 0.427 & 0.411 & 0.399 & 0.391 & 0.472 & 0.430 & 0.462 & 0.425 & 0.406 & 0.398 & 0.429 & 0.413 & 0.404 & 0.403 & 0.414 & 0.404 & 0.419 & 0.411 \\
 & & {\textcolor{blue}{-}} & {\textcolor{blue}{-}} & {\textcolor{blue}{-0.79\%}} & {\textcolor{blue}{0.00\%}} & {\textcolor{blue}{-}} & {\textcolor{blue}{-}} & {\textcolor{red}{+5.97\%}} & {\textcolor{red}{+4.61\%}} & {\textcolor{blue}{-0.91\%}} & {\textcolor{blue}{-0.47\%}} & {\textcolor{red}{+17.24\%}} & {\textcolor{red}{+9.43\%}} & {\textcolor{red}{+14.77\%}} & {\textcolor{red}{+7.95\%}} & {\textcolor{blue}{-}} & {\textcolor{blue}{-}} & {\textcolor{red}{+5.73\%}} & {\textcolor{red}{+3.90\%}} & {\textcolor{blue}{-0.45\%}} & {\textcolor{red}{+1.35\%}} & {\textcolor{red}{+2.09\%}} & {\textcolor{red}{+1.59\%}} & {\textcolor{red}{+3.19\%}} & {\textcolor{red}{+3.36\%}} \\

 & 96-48 & 0.269 & 0.314 & 0.273 & 0.317 & 0.292 & 0.340 & 0.450 & 0.412 & 0.441 & 0.405 & 0.315 & 0.358 & 0.292 & 0.342 & 0.287 & 0.332 & 0.421 & 0.402 & 0.425 & 0.407 & 0.287 & 0.338 & 0.288 & 0.338 \\
 & & {\textcolor{blue}{-}} & {\textcolor{blue}{-}} & {\textcolor{red}{+1.49\%}} & {\textcolor{red}{+0.96\%}} & {\textcolor{blue}{-}} & {\textcolor{blue}{-}} & {\textcolor{red}{+54.30\%}} & {\textcolor{red}{+21.35\%}} & {\textcolor{red}{+51.24\%}} & {\textcolor{red}{+19.38\%}} & {\textcolor{red}{+8.04\%}} & {\textcolor{red}{+5.54\%}} & {\textcolor{red}{+0.20\%}} & {\textcolor{red}{+0.63\%}} & {\textcolor{blue}{-}} & {\textcolor{blue}{-}} & {\textcolor{red}{+46.69\%}} & {\textcolor{red}{+20.86\%}} & {\textcolor{red}{+48.05\%}} & {\textcolor{red}{+22.29\%}} & {\textcolor{red}{+0.06\%}} & {\textcolor{red}{+1.65\%}} & {\textcolor{red}{+0.11\%}} & {\textcolor{red}{+1.81\%}} \\

 & 96-96 & 0.303 & 0.337 & 0.306 & 0.341 & 0.326 & 0.366 & 0.427 & 0.411 & 0.399 & 0.391 & 0.349 & 0.377 & 0.325 & 0.362 & 0.313 & 0.358 & 0.429 & 0.413 & 0.404 & 0.403 & 0.324 & 0.361 & 0.325 & 0.364 \\
 & & {\textcolor{blue}{-}} & {\textcolor{blue}{-}} & {\textcolor{red}{+0.99\%}} & {\textcolor{red}{+1.19\%}} & {\textcolor{blue}{-}} & {\textcolor{blue}{-}} & {\textcolor{red}{+30.96\%}} & {\textcolor{red}{+12.40\%}} & {\textcolor{red}{+22.45\%}} & {\textcolor{red}{+6.94\%}} & {\textcolor{red}{+7.11\%}} & {\textcolor{red}{+3.09\%}} & {\textcolor{blue}{-0.25\%}} & {\textcolor{blue}{-1.21\%}} & {\textcolor{blue}{-}} & {\textcolor{blue}{-}} & {\textcolor{red}{+37.09\%}} & {\textcolor{red}{+15.41\%}} & {\textcolor{red}{+29.08\%}} & {\textcolor{red}{+12.58\%}} & {\textcolor{red}{+3.52\%}} & {\textcolor{red}{+0.93\%}} & {\textcolor{red}{+3.68\%}} & {\textcolor{red}{+1.74\%}} \\

\cmidrule(lr){2-26}
 & AVG & 0.337 & 0.354 & 0.343 & 0.357 & 0.371 & 0.380 & 0.439 & 0.412 & 0.420 & 0.398 & 0.423 & 0.407 & 0.406 & 0.396 & 0.359 & 0.374 & 0.425 & 0.407 & 0.415 & 0.405 & 0.372 & 0.380 & 0.374 & 0.385 \\
 & & {\textcolor{blue}{-}} & {\textcolor{blue}{-}} & {\textcolor{red}{+1.78\%}} & {\textcolor{red}{+0.85\%}} & {\textcolor{blue}{-}} & {\textcolor{blue}{-}} & {\textcolor{red}{+18.23\%}} & {\textcolor{red}{+8.49\%}} & {\textcolor{red}{+13.29\%}} & {\textcolor{red}{+4.98\%}} & {\textcolor{red}{+14.17\%}} & {\textcolor{red}{+7.23\%}} & {\textcolor{red}{+9.51\%}} & {\textcolor{red}{+4.34\%}} & {\textcolor{blue}{-}} & {\textcolor{blue}{-}} & {\textcolor{red}{+18.49\%}} & {\textcolor{red}{+8.91\%}} & {\textcolor{red}{+15.54\%}} & {\textcolor{red}{+8.19\%}} & {\textcolor{red}{+3.60\%}} & {\textcolor{red}{+1.66\%}} & {\textcolor{red}{+4.18\%}} & {\textcolor{red}{+2.79\%}} \\
\midrule
\multirow{10}{*}{ETTm2} & 48-48 & 0.137 & 0.226 & 0.139 & 0.228 & 0.154 & 0.245 & 0.155 & 0.246 & 0.155 & 0.247 & 0.175 & 0.280 & 0.172 & 0.272 & 0.148 & 0.238 & 0.152 & 0.243 & 0.149 & 0.241 & 0.242 & 0.333 & 0.209 & 0.311 \\
 & & {\textcolor{blue}{-}} & {\textcolor{blue}{-}} & {\textcolor{red}{+1.46\%}} & {\textcolor{red}{+0.88\%}} & {\textcolor{blue}{-}} & {\textcolor{blue}{-}} & {\textcolor{red}{+0.70\%}} & {\textcolor{red}{+0.35\%}} & {\textcolor{red}{+0.63\%}} & {\textcolor{red}{+0.65\%}} & {\textcolor{red}{+13.77\%}} & {\textcolor{red}{+14.19\%}} & {\textcolor{red}{+11.68\%}} & {\textcolor{red}{+10.97\%}} & {\textcolor{blue}{-}} & {\textcolor{blue}{-}} & {\textcolor{red}{+2.67\%}} & {\textcolor{red}{+2.24\%}} & {\textcolor{red}{+0.50\%}} & {\textcolor{red}{+1.51\%}} & {\textcolor{red}{+63.78\%}} & {\textcolor{red}{+40.27\%}} & {\textcolor{red}{+41.36\%}} & {\textcolor{red}{+30.89\%}} \\

 & 48-96 & 0.179 & 0.255 & 0.179 & 0.255 & 0.186 & 0.268 & 0.192 & 0.273 & 0.185 & 0.267 & 0.199 & 0.292 & 0.276 & 0.342 & 0.185 & 0.267 & 0.191 & 0.272 & 0.182 & 0.264 & 0.340 & 0.388 & 0.343 & 0.387 \\
 & & {\textcolor{blue}{-}} & {\textcolor{blue}{-}} & {\textcolor{blue}{0.00\%}} & {\textcolor{blue}{0.00\%}} & {\textcolor{blue}{-}} & {\textcolor{blue}{-}} & {\textcolor{red}{+2.82\%}} & {\textcolor{red}{+1.89\%}} & {\textcolor{blue}{-0.48\%}} & {\textcolor{blue}{-0.44\%}} & {\textcolor{red}{+7.10\%}} & {\textcolor{red}{+9.03\%}} & {\textcolor{red}{+48.37\%}} & {\textcolor{red}{+27.65\%}} & {\textcolor{blue}{-}} & {\textcolor{blue}{-}} & {\textcolor{red}{+3.36\%}} & {\textcolor{red}{+1.87\%}} & {\textcolor{blue}{-1.31\%}} & {\textcolor{blue}{-1.06\%}} & {\textcolor{red}{+83.98\%}} & {\textcolor{red}{+45.38\%}} & {\textcolor{red}{+85.69\%}} & {\textcolor{red}{+45.10\%}} \\

 & 96-48 & 0.128 & 0.219 & 0.129 & 0.216 & 0.132 & 0.222 & 0.155 & 0.246 & 0.155 & 0.247 & 0.139 & 0.233 & 0.136 & 0.229 & 0.134 & 0.227 & 0.152 & 0.243 & 0.149 & 0.241 & 0.137 & 0.231 & 0.137 & 0.231 \\
 & & {\textcolor{blue}{-}} & {\textcolor{blue}{-}} & {\textcolor{red}{+0.78\%}} & {\textcolor{blue}{-1.37\%}} & {\textcolor{blue}{-}} & {\textcolor{blue}{-}} & {\textcolor{red}{+17.75\%}} & {\textcolor{red}{+10.80\%}} & {\textcolor{red}{+17.66\%}} & {\textcolor{red}{+11.14\%}} & {\textcolor{red}{+5.71\%}} & {\textcolor{red}{+4.63\%}} & {\textcolor{red}{+3.50\%}} & {\textcolor{red}{+2.96\%}} & {\textcolor{blue}{-}} & {\textcolor{blue}{-}} & {\textcolor{red}{+13.36\%}} & {\textcolor{red}{+7.08\%}} & {\textcolor{red}{+10.97\%}} & {\textcolor{red}{+6.32\%}} & {\textcolor{red}{+2.20\%}} & {\textcolor{red}{+1.80\%}} & {\textcolor{red}{+2.49\%}} & {\textcolor{red}{+1.93\%}} \\

 & 96-96 & 0.168 & 0.249 & 0.171 & 0.247 & 0.173 & 0.253 & 0.192 & 0.273 & 0.185 & 0.267 & 0.178 & 0.259 & 0.174 & 0.256 & 0.173 & 0.252 & 0.191 & 0.272 & 0.182 & 0.264 & 0.177 & 0.259 & 0.176 & 0.259 \\
 & & {\textcolor{blue}{-}} & {\textcolor{blue}{-}} & {\textcolor{red}{+1.79\%}} & {\textcolor{blue}{-0.80\%}} & {\textcolor{blue}{-}} & {\textcolor{blue}{-}} & {\textcolor{red}{+10.71\%}} & {\textcolor{red}{+7.88\%}} & {\textcolor{red}{+7.15\%}} & {\textcolor{red}{+5.42\%}} & {\textcolor{red}{+2.93\%}} & {\textcolor{red}{+2.28\%}} & {\textcolor{red}{+0.86\%}} & {\textcolor{red}{+1.00\%}} & {\textcolor{blue}{-}} & {\textcolor{blue}{-}} & {\textcolor{red}{+10.48\%}} & {\textcolor{red}{+7.80\%}} & {\textcolor{red}{+5.49\%}} & {\textcolor{red}{+4.71\%}} & {\textcolor{red}{+2.11\%}} & {\textcolor{red}{+2.65\%}} & {\textcolor{red}{+1.97\%}} & {\textcolor{red}{+2.62\%}} \\

\cmidrule(lr){2-26}
 & AVG & 0.153 & 0.237 & 0.155 & 0.237 & 0.161 & 0.247 & 0.173 & 0.260 & 0.170 & 0.257 & 0.173 & 0.266 & 0.190 & 0.275 & 0.160 & 0.246 & 0.171 & 0.257 & 0.166 & 0.253 & 0.224 & 0.303 & 0.217 & 0.297 \\
 & & {\textcolor{blue}{-}} & {\textcolor{blue}{-}} & {\textcolor{red}{+0.98\%}} & {\textcolor{blue}{-0.32\%}} & {\textcolor{blue}{-}} & {\textcolor{blue}{-}} & {\textcolor{red}{+7.48\%}} & {\textcolor{red}{+5.04\%}} & {\textcolor{red}{+5.54\%}} & {\textcolor{red}{+3.94\%}} & {\textcolor{red}{+7.29\%}} & {\textcolor{red}{+7.59\%}} & {\textcolor{red}{+17.70\%}} & {\textcolor{red}{+11.14\%}} & {\textcolor{blue}{-}} & {\textcolor{blue}{-}} & {\textcolor{red}{+7.22\%}} & {\textcolor{red}{+4.68\%}} & {\textcolor{red}{+3.52\%}} & {\textcolor{red}{+2.74\%}} & {\textcolor{red}{+40.05\%}} & {\textcolor{red}{+23.14\%}} & {\textcolor{red}{+35.38\%}} & {\textcolor{red}{+20.82\%}} \\
\midrule
\multirow{10}{*}{Weather} & 48-48 & 0.140 & 0.176 & 0.142 & 0.176 & 0.150 & 0.183 & 0.166 & 0.197 & 0.153 & 0.185 & 0.150 & 0.199 & 0.157 & 0.204 & 0.138 & 0.179 & 0.144 & 0.181 & 0.141 & 0.183 & 0.149 & 0.206 & 0.148 & 0.201 \\
 & & {\textcolor{blue}{-}} & {\textcolor{blue}{-}} & {\textcolor{red}{+1.92\%}} & {\textcolor{red}{+0.27\%}} & {\textcolor{blue}{-}} & {\textcolor{blue}{-}} & {\textcolor{red}{+11.05\%}} & {\textcolor{red}{+7.87\%}} & {\textcolor{red}{+2.10\%}} & {\textcolor{red}{+1.10\%}} & {\textcolor{blue}{-0.06\%}} & {\textcolor{red}{+9.01\%}} & {\textcolor{red}{+5.05\%}} & {\textcolor{red}{+11.76\%}} & {\textcolor{blue}{-}} & {\textcolor{blue}{-}} & {\textcolor{red}{+4.97\%}} & {\textcolor{red}{+1.15\%}} & {\textcolor{red}{+2.81\%}} & {\textcolor{red}{+2.02\%}} & {\textcolor{red}{+8.40\%}} & {\textcolor{red}{+14.55\%}} & {\textcolor{red}{+7.36\%}} & {\textcolor{red}{+12.22\%}} \\

 & 48-96 & 0.172 & 0.214 & 0.178 & 0.218 & 0.192 & 0.232 & 0.202 & 0.239 & 0.190 & 0.228 & 0.187 & 0.242 & 0.193 & 0.246 & 0.175 & 0.223 & 0.179 & 0.224 & 0.174 & 0.223 & 0.195 & 0.257 & 0.196 & 0.258 \\
 & & {\textcolor{blue}{-}} & {\textcolor{blue}{-}} & {\textcolor{red}{+3.16\%}} & {\textcolor{red}{+2.04\%}} & {\textcolor{blue}{-}} & {\textcolor{blue}{-}} & {\textcolor{red}{+5.20\%}} & {\textcolor{red}{+3.02\%}} & {\textcolor{blue}{-1.11\%}} & {\textcolor{blue}{-1.40\%}} & {\textcolor{blue}{-2.74\%}} & {\textcolor{red}{+4.57\%}} & {\textcolor{red}{+0.72\%}} & {\textcolor{red}{+6.43\%}} & {\textcolor{blue}{-}} & {\textcolor{blue}{-}} & {\textcolor{red}{+2.43\%}} & {\textcolor{red}{+0.15\%}} & {\textcolor{blue}{-0.50\%}} & {\textcolor{blue}{-0.13\%}} & {\textcolor{red}{+11.43\%}} & {\textcolor{red}{+15.18\%}} & {\textcolor{red}{+12.25\%}} & {\textcolor{red}{+15.53\%}} \\

 & 96-48 & 0.119 & 0.151 & 0.119 & 0.153 & 0.130 & 0.168 & 0.166 & 0.197 & 0.153 & 0.185 & 0.130 & 0.168 & 0.134 & 0.173 & 0.127 & 0.167 & 0.144 & 0.181 & 0.141 & 0.183 & 0.125 & 0.168 & 0.126 & 0.168 \\
 & & {\textcolor{blue}{-}} & {\textcolor{blue}{-}} & {\textcolor{blue}{-0.13\%}} & {\textcolor{red}{+1.37\%}} & {\textcolor{blue}{-}} & {\textcolor{blue}{-}} & {\textcolor{red}{+27.97\%}} & {\textcolor{red}{+17.08\%}} & {\textcolor{red}{+17.66\%}} & {\textcolor{red}{+9.73\%}} & {\textcolor{red}{+0.08\%}} & {\textcolor{red}{+0.14\%}} & {\textcolor{red}{+3.07\%}} & {\textcolor{red}{+2.89\%}} & {\textcolor{blue}{-}} & {\textcolor{blue}{-}} & {\textcolor{red}{+13.73\%}} & {\textcolor{red}{+8.99\%}} & {\textcolor{red}{+11.39\%}} & {\textcolor{red}{+9.92\%}} & {\textcolor{blue}{-1.36\%}} & {\textcolor{red}{+0.69\%}} & {\textcolor{blue}{-0.97\%}} & {\textcolor{red}{+0.86\%}} \\

 & 96-96 & 0.149 & 0.187 & 0.150 & 0.190 & 0.168 & 0.207 & 0.202 & 0.239 & 0.190 & 0.228 & 0.164 & 0.208 & 0.169 & 0.213 & 0.155 & 0.197 & 0.179 & 0.224 & 0.174 & 0.223 & 0.162 & 0.210 & 0.160 & 0.205 \\
 & & {\textcolor{blue}{-}} & {\textcolor{blue}{-}} & {\textcolor{red}{+0.67\%}} & {\textcolor{red}{+1.29\%}} & {\textcolor{blue}{-}} & {\textcolor{blue}{-}} & {\textcolor{red}{+20.27\%}} & {\textcolor{red}{+15.26\%}} & {\textcolor{red}{+13.05\%}} & {\textcolor{red}{+10.31\%}} & {\textcolor{blue}{-2.20\%}} & {\textcolor{red}{+0.43\%}} & {\textcolor{red}{+0.66\%}} & {\textcolor{red}{+2.72\%}} & {\textcolor{blue}{-}} & {\textcolor{blue}{-}} & {\textcolor{red}{+15.59\%}} & {\textcolor{red}{+13.59\%}} & {\textcolor{red}{+12.29\%}} & {\textcolor{red}{+13.27\%}} & {\textcolor{red}{+4.78\%}} & {\textcolor{red}{+6.43\%}} & {\textcolor{red}{+2.97\%}} & {\textcolor{red}{+4.19\%}} \\

\cmidrule(lr){2-26}
 & AVG & 0.145 & 0.182 & 0.147 & 0.184 & 0.160 & 0.197 & 0.184 & 0.218 & 0.171 & 0.206 & 0.158 & 0.204 & 0.163 & 0.209 & 0.149 & 0.192 & 0.162 & 0.203 & 0.158 & 0.203 & 0.158 & 0.210 & 0.157 & 0.208 \\
 & & {\textcolor{blue}{-}} & {\textcolor{blue}{-}} & {\textcolor{red}{+1.55\%}} & {\textcolor{red}{+1.28\%}} & {\textcolor{blue}{-}} & {\textcolor{blue}{-}} & {\textcolor{red}{+15.15\%}} & {\textcolor{red}{+10.35\%}} & {\textcolor{red}{+7.17\%}} & {\textcolor{red}{+4.62\%}} & {\textcolor{blue}{-1.40\%}} & {\textcolor{red}{+3.57\%}} & {\textcolor{red}{+2.20\%}} & {\textcolor{red}{+5.93\%}} & {\textcolor{blue}{-}} & {\textcolor{blue}{-}} & {\textcolor{red}{+8.86\%}} & {\textcolor{red}{+5.76\%}} & {\textcolor{red}{+6.14\%}} & {\textcolor{red}{+6.00\%}} & {\textcolor{red}{+6.26\%}} & {\textcolor{red}{+9.63\%}} & {\textcolor{red}{+5.87\%}} & {\textcolor{red}{+8.65\%}} \\
\midrule
\multirow{10}{*}{Solar} & 48-48 & 0.325 & 0.361 & 0.297 & 0.330 & 0.611 & 0.421 & 0.570 & 0.417 & 0.363 & 0.359 & 0.342 & 0.384 & 0.331 & 0.383 & 0.246 & 0.270 & 0.249 & 0.281 & 0.262 & 0.276 & 0.237 & 0.267 & 0.245 & 0.259 \\
 & & {\textcolor{blue}{-}} & {\textcolor{blue}{-}} & {\textcolor{blue}{-8.90\%}} & {\textcolor{blue}{-8.81\%}} & {\textcolor{blue}{-}} & {\textcolor{blue}{-}} & {\textcolor{blue}{-6.82\%}} & {\textcolor{blue}{-0.94\%}} & {\textcolor{blue}{-40.61\%}} & {\textcolor{blue}{-14.82\%}} & {\textcolor{blue}{-44.06\%}} & {\textcolor{blue}{-8.73\%}} & {\textcolor{blue}{-45.88\%}} & {\textcolor{blue}{-9.00\%}} & {\textcolor{blue}{-}} & {\textcolor{blue}{-}} & {\textcolor{red}{+1.49\%}} & {\textcolor{red}{+4.24\%}} & {\textcolor{red}{+6.58\%}} & {\textcolor{red}{+2.55\%}} & {\textcolor{blue}{-3.48\%}} & {\textcolor{blue}{-1.06\%}} & {\textcolor{blue}{-0.33\%}} & {\textcolor{blue}{-3.73\%}} \\

 & 48-96 & 0.344 & 0.372 & 0.350 & 0.373 & 0.597 & 0.424 & 0.600 & 0.445 & 0.401 & 0.388 & 0.417 & 0.432 & 0.376 & 0.439 & 0.318 & 0.314 & 0.288 & 0.309 & 0.318 & 0.308 & 0.279 & 0.295 & 0.279 & 0.289 \\
 & & {\textcolor{blue}{-}} & {\textcolor{blue}{-}} & {\textcolor{red}{+1.78\%}} & {\textcolor{red}{+0.40\%}} & {\textcolor{blue}{-}} & {\textcolor{blue}{-}} & {\textcolor{red}{+0.64\%}} & {\textcolor{red}{+5.03\%}} & {\textcolor{blue}{-32.81\%}} & {\textcolor{blue}{-8.60\%}} & {\textcolor{blue}{-30.07\%}} & {\textcolor{red}{+1.77\%}} & {\textcolor{blue}{-37.03\%}} & {\textcolor{red}{+3.49\%}} & {\textcolor{blue}{-}} & {\textcolor{blue}{-}} & {\textcolor{blue}{-9.46\%}} & {\textcolor{blue}{-1.74\%}} & {\textcolor{red}{+0.02\%}} & {\textcolor{blue}{-1.93\%}} & {\textcolor{blue}{-12.24\%}} & {\textcolor{blue}{-6.21\%}} & {\textcolor{blue}{-12.31\%}} & {\textcolor{blue}{-8.16\%}} \\

 & 96-48 & 0.139 & 0.198 & 0.144 & 0.210 & 0.203 & 0.245 & 0.570 & 0.417 & 0.363 & 0.359 & 0.277 & 0.326 & 0.302 & 0.355 & 0.179 & 0.197 & 0.249 & 0.281 & 0.262 & 0.276 & 0.176 & 0.207 & 0.186 & 0.198 \\
 & & {\textcolor{blue}{-}} & {\textcolor{blue}{-}} & {\textcolor{red}{+3.84\%}} & {\textcolor{red}{+5.97\%}} & {\textcolor{blue}{-}} & {\textcolor{blue}{-}} & {\textcolor{red}{+180.45\%}} & {\textcolor{red}{+70.05\%}} & {\textcolor{red}{+78.75\%}} & {\textcolor{red}{+46.22\%}} & {\textcolor{red}{+36.47\%}} & {\textcolor{red}{+33.07\%}} & {\textcolor{red}{+48.85\%}} & {\textcolor{red}{+44.93\%}} & {\textcolor{blue}{-}} & {\textcolor{blue}{-}} & {\textcolor{red}{+38.93\%}} & {\textcolor{red}{+42.34\%}} & {\textcolor{red}{+45.91\%}} & {\textcolor{red}{+40.02\%}} & {\textcolor{blue}{-1.75\%}} & {\textcolor{red}{+4.79\%}} & {\textcolor{red}{+3.63\%}} & {\textcolor{red}{+0.56\%}} \\

 & 96-96 & 0.163 & 0.214 & 0.168 & 0.225 & 0.184 & 0.219 & 0.600 & 0.445 & 0.401 & 0.388 & 0.293 & 0.341 & 0.291 & 0.339 & 0.187 & 0.218 & 0.288 & 0.309 & 0.318 & 0.308 & 0.209 & 0.232 & 0.226 & 0.226 \\
 & & {\textcolor{blue}{-}} & {\textcolor{blue}{-}} & {\textcolor{red}{+3.30\%}} & {\textcolor{red}{+5.29\%}} & {\textcolor{blue}{-}} & {\textcolor{blue}{-}} & {\textcolor{red}{+226.34\%}} & {\textcolor{red}{+103.37\%}} & {\textcolor{red}{+117.88\%}} & {\textcolor{red}{+77.00\%}} & {\textcolor{red}{+59.25\%}} & {\textcolor{red}{+55.82\%}} & {\textcolor{red}{+58.21\%}} & {\textcolor{red}{+54.89\%}} & {\textcolor{blue}{-}} & {\textcolor{blue}{-}} & {\textcolor{red}{+54.13\%}} & {\textcolor{red}{+41.67\%}} & {\textcolor{red}{+70.27\%}} & {\textcolor{red}{+41.40\%}} & {\textcolor{red}{+11.97\%}} & {\textcolor{red}{+6.41\%}} & {\textcolor{red}{+20.90\%}} & {\textcolor{red}{+3.60\%}} \\

\cmidrule(lr){2-26}
 & AVG & 0.243 & 0.286 & 0.240 & 0.285 & 0.399 & 0.327 & 0.585 & 0.431 & 0.382 & 0.373 & 0.332 & 0.371 & 0.325 & 0.379 & 0.233 & 0.250 & 0.269 & 0.295 & 0.290 & 0.292 & 0.226 & 0.250 & 0.234 & 0.243 \\
 & & {\textcolor{blue}{-}} & {\textcolor{blue}{-}} & {\textcolor{blue}{-1.25\%}} & {\textcolor{blue}{-0.63\%}} & {\textcolor{blue}{-}} & {\textcolor{blue}{-}} & {\textcolor{red}{+46.71\%}} & {\textcolor{red}{+31.74\%}} & {\textcolor{blue}{-4.21\%}} & {\textcolor{red}{+13.99\%}} & {\textcolor{blue}{-16.66\%}} & {\textcolor{red}{+13.30\%}} & {\textcolor{blue}{-18.50\%}} & {\textcolor{red}{+15.83\%}} & {\textcolor{blue}{-}} & {\textcolor{blue}{-}} & {\textcolor{red}{+15.54\%}} & {\textcolor{red}{+18.05\%}} & {\textcolor{red}{+24.72\%}} & {\textcolor{red}{+17.02\%}} & {\textcolor{blue}{-3.04\%}} & {\textcolor{red}{+0.11\%}} & {\textcolor{red}{+0.60\%}} & {\textcolor{blue}{-2.68\%}} \\
\bottomrule
\end{tabular}%
}
\label{tab:multilen_other}
\end{table*}



\end{document}